\documentclass[letterpaper, 10 pt, journal, twoside]{IEEEtran}

\usepackage{cite}
\usepackage{amsmath,amssymb,amsfonts}
\usepackage{graphicx}
\usepackage{booktabs}
\usepackage{multirow}
\usepackage[table,dvipsnames]{xcolor}
\usepackage{rotating}
\usepackage{tikz}
\usetikzlibrary{arrows.meta,positioning,shapes,automata,backgrounds}
\usepackage{pgfplots}
\usepackage{siunitx}
\usepackage{afterpage}
\usepackage{acro}
\usepackage{url}
\usepackage[normalem]{ulem} % for \sout and \deleted
\DeclareAcronym{lru}{ 
    short = {LRU}, 
    long  = {Least Recently Used},
}

\DeclareAcronym{vram}{ 
    short = {VRAM}, 
    long  = {Video Random Access Memory},
}

\DeclareAcronym{slam}{ 
    short = {SLAM}, 
    long  = {Simultaneous Localization and Mapping},
}

\DeclareAcronym{ar}{ 
    short = {AR}, 
    long  = {Augmented Reality},
}

\DeclareAcronym{lpips}{ 
    short = {LPIPS}, 
    long  = {Learned Perceptual Image Patch Similarity},
}

\DeclareAcronym{psnr}{ 
    short = {PSNR}, 
    long  = {Peak Signal-to-Noise Ratio},
}

\DeclareAcronym{ssim}{ 
    short = {SSIM}, 
    long  = {Structural Similarity Index Measure},
}

\DeclareAcronym{ate}{ 
    short = {ATE}, 
    long  = {Absolute Tracking Error},
}

\DeclareAcronym{3dgs}{ 
    short = {3DGS}, 
    long  = {3D Gaussian Splatting},
}

\DeclareAcronym{log}{ 
    short = {LoG}, 
    long  = {Laplacian of Gaussian},
}

\DeclareAcronym{lod}{ 
    short = {LoD}, 
    long  = {Level of Detail},
}

\pgfplotsset{width=10cm,compat=1.18}

\definecolor{colorFst}{HTML}{bde6cd}       % first
\definecolor{colorSnd}{HTML}{e4eebc}       % second

\newcommand{\fs}{\cellcolor{colorFst}\bf}   % first
\newcommand{\fscap}[1]{\colorbox{colorFst}{\textbf{#1}}}   % for captions/text
\newcommand{\nd}{\cellcolor{colorSnd}\underline}      % second
\newcommand{\ndcap}[1]{\colorbox{colorSnd}{\underline{#1}}}   % for captions/text

\title{DiskChunGS: Large-Scale 3D Gaussian SLAM Through Chunk-Based Memory Management
}

\author{Casimir Feldmann$^{1}$, Maximum Wilder-Smith$^{1}$, Vaishakh Patil$^{1}$, Member, IEEE, \\
Michael Oechsle$^{2}$, Michael Niemeyer$^{2}$, Keisuke Tateno$^{2}$, and Marco Hutter$^{1}$, Member, IEEE%
\thanks{Received 21 October 2025; accepted 31 January 2026.}
\thanks{This article was recommended for publication by Associate Editor Y. Li and Editor J. Civera upon evaluation of the reviewers' comments.
This work was supported by NCCR Automation, Swiss Federal Railways
(SBB) via ETH Mobility Initiative, ETHAR, and the ETH Augmented Reality Research.}
\thanks{$^{1}$Casimir Feldmann, Maximum Wilder-Smith, Vaishakh Patil, and Marco Hutter are with Robotic Systems Lab, ETH Zurich, 8092 Zürich, Switzerland
        {\tt\footnotesize casimir.feldmann@gmail.com}, and {\tt\footnotesize \{mwilder, patilv, mahutter\}@ethz.ch}}%
\thanks{$^{2} $Michael Oechsle, Michael Niemeyer, and Keisuke Tateno are with Google, 8002 Zürich, Switzerland
{\tt\footnotesize \{michaeloechsle, mniemeyer, ktateno\}@google.com}}%}%
\thanks{Digital Object Identifier 10.1109/LRA.2026.3668704}
}

\begin{document}
\maketitle

%%%%%%%%%%%%%%%%%%%%%%%%%%%%%%%%%%%%%%%%%%%%%%%%%%%%%%%%%%%%%%%%%%%%%%%%%%%%%%%%
\begin{abstract}
Recent advances in 3D Gaussian Splatting (3DGS) have demonstrated impressive results for novel view synthesis with real-time rendering capabilities. However, integrating 3DGS with SLAM systems faces a fundamental scalability limitation: methods are constrained by GPU memory capacity, restricting reconstruction to small-scale environments. We present DiskChunGS, a scalable 3DGS SLAM system that overcomes this bottleneck through an out-of-core approach that partitions scenes into spatial chunks and maintains only active regions in GPU memory while storing inactive areas on disk. Our architecture integrates seamlessly with existing SLAM frameworks for pose estimation and loop closure, enabling globally consistent reconstruction at scale. We validate DiskChunGS on indoor scenes (Replica, TUM-RGBD), urban driving scenarios (KITTI), and resource-constrained Nvidia Jetson platforms. Our method uniquely completes all 11 KITTI sequences without memory failures while achieving superior visual quality, demonstrating that algorithmic innovation can overcome the memory constraints that have limited previous 3DGS SLAM methods. Project page and code: \url{https://rffr.leggedrobotics.com/works/diskchungs}
\end{abstract} 


\section{Introduction}

\IEEEPARstart{R}{ecent} advances in neural representations for 3D scene reconstruction have revolutionized novel view synthesis, with \ac{3dgs}~\cite{kerbl3Dgaussians} emerging as an exceptionally efficient and high-quality approach. Unlike volume-based methods~\cite{mildenhall2020nerfrepresentingscenesneural, M_ller_2022} that struggle with rendering speed due to expensive ray marching, \ac{3dgs} provides real-time rendering capabilities while maintaining impressive visual fidelity. However, extending \ac{3dgs} to large-scale environments for \ac{slam} applications introduces a fundamental bottleneck: existing methods require the entire scene representation to fit within GPU \ac{vram}, severely limiting the size of environments that can be reconstructed.

Our work presents a novel \ac{3dgs}-based \ac{slam} system that overcomes this memory limitation through an out-of-core architecture inspired by virtual memory systems. Crucially, our approach does not achieve scalability through Gaussian efficiency or representation compression. Instead, the key insight is to divide the scene into spatial regions, like tiles on a map, and maintain only the currently relevant areas in GPU memory,  storing the remainder on disk. We refer to these spatial regions as ``chunks'' and dynamically load them based on which parts of the scene are visible from the current camera position. This approach allows our system to reconstruct substantially larger environments without compromising visual quality. 

Beyond memory constraints, large-scale SLAM faces additional challenges, including pose drift accumulation and loop closure detection in expansive environments. To ensure robust localization and global consistency, our system integrates with the proven ORB-SLAM3~\cite{Campos_2021} framework and introduces chunk-aware loop closure mechanisms that operate seamlessly within our out-of-core architecture.

Our approach enables practical, real-world deployment advantages. The spatial partitioning enables efficient serialization and incremental map updates while scaling across hardware platforms, from memory-constrained mobile robots to high-end systems. This opens up applications in autonomous navigation, AR/VR, digital twins, and cultural heritage preservation, which require photorealistic reconstruction at scale.

\begin{figure*}[htb]
    \centering
    \newcommand{\plotscale}{0.8}
    \newcommand{\sw}{6.5cm}   
    \newcommand{\sh}{5cm}
    \newcommand{\plotlinewidth}{1.2pt}
    \newcommand{\marksize}{1.2pt}
    % First graph
    \begin{tikzpicture}[scale=\plotscale]
    \begin{axis}[
        width=\sw,
        height=\sh, 
        xlabel={FPS},
        ylabel={LPIPS},
        xmin=0, xmax=10,
        ymin=0.2, ymax=0.6,
        xtick={0,2,4,6,8,10},
        ytick={0.2,0.3,0.4,0.5,0.6},
        ymajorgrids=true,
        grid style=dashed,
        title={Scene 06},
        name=plot1,
    ]
   \addplot[color=violet, mark=triangle*, line width=\plotlinewidth, mark size=\marksize]
        coordinates {(1.1506123326831,0.40006709642891)(2.22994703420083,0.44548579444136)(3.2877792260729,0.415451820713081)(4.27620144365493,0.487713430655425)(6.050930035091,0.512805094971211)(7.81163529241484,0.546427108448922)(9.16291259114019,0.566145407027488)};
    \addplot[color=blue, mark=square*, line width=\plotlinewidth, mark size=\marksize]
        coordinates {(1.7116533507302,0.58719145417859)(1.56552064122847,0.519733579042586)(1.29989485917622,0.421683311677582)(1.02223689649931,0.348051149034973)};
    \addplot[color=red, mark=diamond*, line width=\plotlinewidth, mark size=\marksize]
        coordinates {(7.61432701705317,0.52986058689886)(5.08505184320674,0.44874951626453)(2.76840210120647,0.399207733276874)(1.5976396056849,0.487984521657482)(1.01864400653429,0.409940611002809)};
    \addplot[color=orange, mark=*, line width=\plotlinewidth, mark size=\marksize]
        coordinates {(1.13981550517638,0.312315926977566)(2.17556041428519,0.322576713653994)(3.10918140421476,0.329563178131964)(3.97968589036887,0.32717931895897)(5.53531414793115,0.340230494905323)(6.7729005238694,0.347705741584924)(7.86444862072143,0.362993266762567)(8.88984345478211,0.36684842771954)(9.66734159520795,0.374872990010321)};
    \end{axis}
    \end{tikzpicture}
    \hfill
    % Second graph
    \begin{tikzpicture}[scale=\plotscale]
    \begin{axis}[
        width=\sw,
        height=\sh,
        xlabel={FPS},
        ylabel={LPIPS},
        xmin=0, xmax=15,
        ymin=0.2, ymax=0.6,
        xtick={0,3,6,9,12,15},
        ytick={0.2,0.3,0.4,0.5,0.6},
        ymajorgrids=true,
        grid style=dashed,
        title={Scene 07},
        name=plot2,
    ]
    \addplot[color=violet, mark=triangle*, line width=\plotlinewidth, mark size=\marksize]
        coordinates {(1.31597379431222,0.344452907574058)(2.13804633055128,0.363005116879669)(4.05961625749102,0.380568213131512)(5.84543522477318,0.388739980222635)(7.33151011129113,0.411900805977666)(9.948711275929,0.429883122511824)(12.161244180331,0.449194397658895)};
    \addplot[color=blue, mark=square*, line width=\plotlinewidth, mark size=\marksize]
        coordinates {((1.96502034972535,0.522558650380177)(1.72894255554858,0.455716737061451)(1.44019942602763,0.362289987547716)(1.05812574510614,0.292367740197385)};
    \addplot[color=red, mark=diamond*, line width=\plotlinewidth, mark size=\marksize]
        coordinates {(12.9771277206148,0.463199627013301)(9.13248872250255,0.386242806762952)(5.6391956535486,0.397585713437625)(3.0803814859093,0.355700911032005)(1.72452135413581,0.383523043170571)(1.0973315229425,0.346663512933096)};
    \addplot[color=orange, mark=*, line width=\plotlinewidth, mark size=\marksize]
        coordinates {(1.29377020854965,0.25255251533288)(2.07816191580481,0.252443373064791)(3.83852935167553,0.257525827233673)(5.32700865965495,0.249761670234081)(6.59556318082267,0.263872426340575)(8.7985679351735,0.271902350146459)(10.4492017891767,0.282086686840218)(12.0171842460281,0.299214791506529)(13.570444287225,0.309673493820078)(14.4509049862841,0.315866609966313)};
    \end{axis}
    \end{tikzpicture}
    \hfill
    % Third graph
    \begin{tikzpicture}[scale=\plotscale]
    \begin{axis}[
        width=\sw,
        height=\sh, 
        xlabel={FPS},
        ylabel={LPIPS},
        xmin=0, xmax=15,
        ymin=0.2, ymax=0.6,
        xtick={0,3,6,9,12,15},
        ytick={0.2,0.3,0.4,0.5,0.6},
        ymajorgrids=true,
        grid style=dashed,
        title={Scene 10},
        name=plot3,
        legend to name=mylegend,
        legend style={
            draw=none,
            fill=none,
            legend columns=1,
            font=\footnotesize
        },
    ]
     \addplot[color=violet, mark=triangle*, line width=\plotlinewidth, mark size=\marksize]
        coordinates {(1.01888621399444,0.353158508609673)(1.66555606482217,0.377425085260012)(3.18364038276847,0.407068703667905)(4.56460647467195,0.424483202044215)(5.86946237484019,0.444759027584209)(7.89875908587484,0.450175782909103)(9.93834255588352,0.473410697577895)(11.7030067275236,0.466820371910297)(14.0168482282284,0.483076383405482)};
    \addplot[color=blue, mark=square*, line width=\plotlinewidth, mark size=\marksize]
        coordinates {(1.81152845809327,0.588977468617731)(1.63446011416153,0.532126054415275)(1.23264231507444,0.4297219776731)(0.963824785239087,0.340123970148175)};
    \addplot[color=red, mark=diamond*, line width=\plotlinewidth, mark size=\marksize]
        coordinates {(12.9688929923199,0.44834937136444)(8.77880593962372,0.388461386649766)(5.97791884768655,0.342936652127877)(3.35811472034458,0.320252002460723)(1.78779128910024,0.312328735646469)(1.2283030228944,0.33953270028475)};
    \addplot[color=orange, mark=*, line width=\plotlinewidth, mark size=\marksize]
        coordinates {(1.02417915579667,0.211350099306755)(1.6808428006116,0.220514156935959)(3.23307220195906,0.237560950667786)(4.65809540911924,0.253578303955999)(6.02183705640972,0.265631562719742)(8.36088969334122,0.287544342536687)(10.6669556188921,0.306332259585983)(12.648309591863,0.323252441784538)(14.254820645633,0.33629153225873)};
    \legend{CaRtGS, GigaSLAM, On-The-Fly, \textbf{Ours}}
    \end{axis}
    
    % Place legend to the right
    \node[right=0mm, anchor=west] at (plot3.east) {\pgfplotslegendfromname{mylegend}};
    \end{tikzpicture}
    \caption{\textbf{Pareto curves for KITTI~\cite{Geiger2012CVPR} scenes.} With more iterations and, as a consequence, more processing time, \ac{3dgs} \ac{slam} methods can optimize for longer, achieving higher reconstruction quality. Our method achieves superior visual quality in less time than competing methods across all three scenes.}
    \label{fig:pareto_comparison}
\end{figure*}

We validate our approach through extensive evaluations on diverse datasets spanning different scales and environments, including Replica~\cite{straub2019replicadatasetdigitalreplica} (indoor scenes), TUM~\cite{sturm12iros} (office environments), and KITTI~\cite{Geiger2012CVPR} (urban driving scenarios). Our evaluation demonstrates robust performance across settings, from highly detailed indoor spaces to challenging outdoor environments. We also validate real-world deployment by demonstrating efficient online processing on indoor datasets and successful large-scale reconstruction on the compute-constrained Nvidia Jetson Orin platform.

Our contributions include:
\begin{itemize}
    \item A novel out-of-core chunk-based architecture that enables large-scale \ac{3dgs} \ac{slam} by partitioning scenes into spatial regions and dynamically managing them between disk and \ac{vram}, overcoming the fundamental memory limitations of previous methods.
    \item Comprehensive evaluation demonstrating state-of-the-art performance across indoor and outdoor datasets, with our method uniquely completing all 11 KITTI~\cite{Geiger2012CVPR} sequences without memory failures while achieving superior quality.
    \item Production-ready deployment validated on edge hardware (Jetson Orin) and integrated with ROS for robotic platforms, bridging the gap between research prototypes and real-world applications.
\end{itemize} 
\section{Related Work}

\subsection{Neural Large-Scale Reconstructions}

Recent neural approaches have pursued large-scale scene representation through spatial partitioning. Block-NeRF~\cite{tancik2022blocknerfscalablelargescene} partitions city-scale scenes into individually trained NeRF models with blending, while Mega-NeRF~\cite{Turki_2022_CVPR} uses geometry-aware pixel-data partitioning for parallel training of spatial cell submodules. For 3DGS, CityGaussian~\cite{liu2024citygaussianrealtimehighqualitylargescale} enables real-time rendering through divide-and-conquer training and block-wise Level-of-Detail strategies that dynamically adjust detail based on viewing distance. However, these approaches sacrifice real-time performance or require substantial computational resources for training and inference.

\subsection{Gaussian Splatting SLAM}

The integration of 3DGS into SLAM represents a significant advancement in real-time photorealistic reconstruction. Early methods like GS-SLAM~\cite{yan2024gsslamdensevisualslam} first demonstrated this potential. Photo-SLAM~\cite{huang2024photoslamrealtimesimultaneouslocalization} advanced the field by combining ORB-SLAM3 localization~\cite{Campos_2021} with hyper primitives that merge explicit geometry with implicit photometric representations. CaRtGS~\cite{Feng_2025} extends Photo-SLAM by improving rendering efficiency through splat-wise backpropagation, adaptive keyframe optimization, and opacity regularization for model size control.

Several recent methods employ submaps or hierarchical representations to improve reconstruction quality: Gaussian-SLAM~\cite{yugay2024gaussianslamphotorealisticdenseslam} uses submaps to prevent catastrophic forgetting, GLC-SLAM~\cite{xu2024glcslamgaussiansplattingslam} employs hierarchical loop closure for global consistency, and MemGS~\cite{bai2025memgsmemoryefficientgaussiansplatting} reduces memory through voxel-based Gaussian merging. However, these have been validated exclusively on room-scale indoor datasets.

For large-scale outdoor environments, GigaSLAM~\cite{deng2025gigaslamlargescalemonocularslam} targets kilometer-scale environments using hierarchical sparse voxel maps with level-of-detail rendering. While they demonstrate results on urban driving scenarios spanning multiple kilometers, achieving high visual quality requires substantial offline post-processing and systems with up to 48 GB VRAM for long sequences~\cite{deng2025gigaslamlargescalemonocularslam}. On-The-Fly~\cite{Meuleman_2025} addresses GPU memory constraints through incremental clustering and anchoring, storing distant content in CPU RAM rather than disk. While this anchor-based approach with progressive merging shifts the memory bottleneck from VRAM to system RAM, it does not fundamentally eliminate scalability limits. System RAM is similarly constrained and expensive, particularly on embedded platforms with unified memory architectures in which the CPU and GPU share the same memory pool, rendering RAM-based approaches ineffective. Additionally, On-The-Fly lacks loop-closure detection to maintain global consistency.

In contrast, our approach scales through algorithmic innovation rather than hardware requirements. By treating reconstruction as an out-of-core problem with disk-based chunk management, we enable large-scale mapping on standard hardware while maintaining full representation fidelity (3rd-degree spherical harmonics) and incorporating robust loop closure for global consistency.

\begin{figure*}[t]
    \centering
    \newcommand{\sz}{1.0}
    \includegraphics[width=\sz\linewidth]{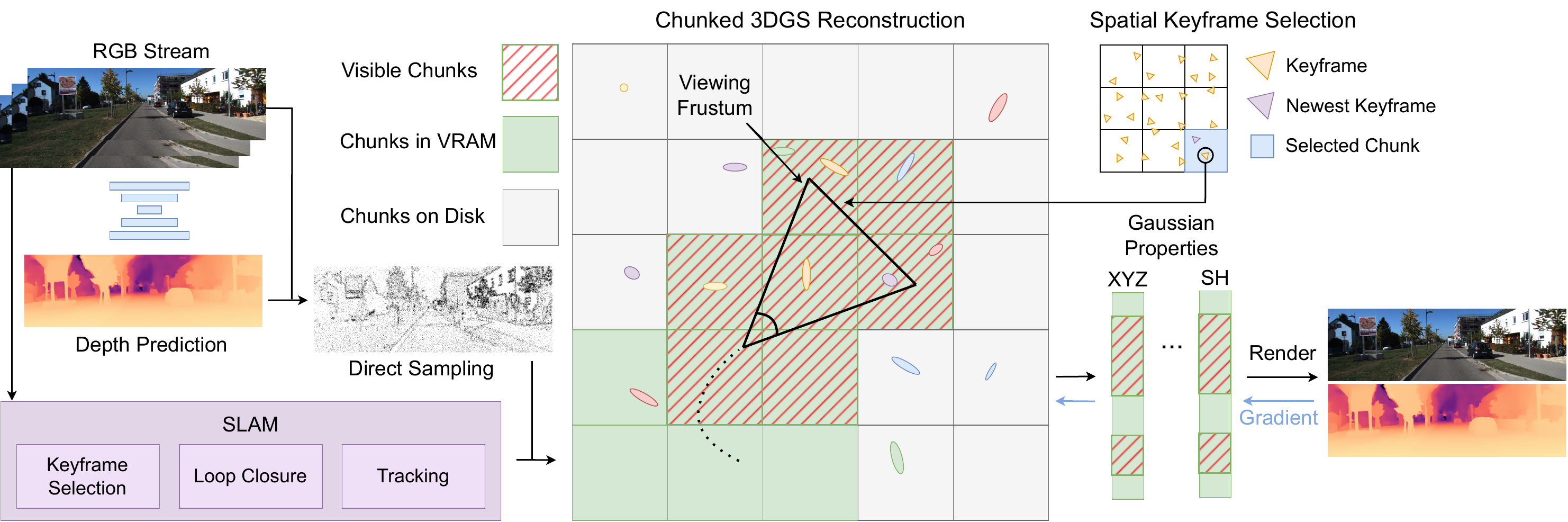}
    \caption{\textbf{Overview of DiskChunGS.} For each \ac{slam} keyframe, we estimate depth and perform direct primitive placement based on image content analysis instead of iterative densification. For optimization of a keyframe, frustum culling is performed to identify visible chunks, which are loaded from disk into \ac{vram}. On the other hand, old chunks are evicted from \ac{vram} to disk to free up memory. The visible subset of Gaussians in \ac{vram} is then rasterized, and image/depth losses are calculated.}
    \label{fig:pipeline}
\end{figure*} 
\section{Method}

Our approach combines robust visual SLAM with chunk-based 3D Gaussian Splatting to enable scalable dense mapping of large-scale environments. As shown in Figure~\ref{fig:pipeline}, the system operates through parallel tracking and mapping threads, where accurate pose estimation guides efficient Gaussian placement and optimization within our hierarchical scene representation.

\subsection{Tracking}
Our system supports two tracking modes for different deployments. In standalone mode, we employ ORB-SLAM3~\cite{Campos_2021} for feature-based visual odometry, sharing co-visibility-optimized keyframes between the tracking and mapping threads. For robotic integration, our ROS wrapper enables external pose input, allowing integration with existing SLAM systems that may achieve superior localization through multi-sensor fusion. The system accepts monocular, stereo, and RGB-D camera inputs. All experiments use the standalone ORB-SLAM3 mode without IMU.

\subsection{Sampling}
The sparse point cloud $P \in \mathbb{R}^3$ from ORB-SLAM3~\cite{Campos_2021} provides one source of Gaussians, with keypoints from each keyframe contributing Gaussians to the scene representation. To efficiently add complementary Gaussians, we simultaneously adopt the direct primitive sampling method from Meuleman et al.~\cite{Meuleman_2025} for each keyframe. This method estimates the probability that each pixel should generate a Gaussian primitive based on the norm of the \ac{log} operator, which identifies high-frequency details and edges in the image. To avoid placing redundant Gaussians, the current scene representation is rendered, and a penalty map is computed from the rendered image's \ac{log} response. The final sampling probability is then given by:

\begin{equation}
P_s(x, y) = \max\left(P_{\text{L}}(x, y) - \tilde{P}(x, y), 0\right)
\end{equation}
 
\noindent where $P_{\text{L}}$ is the \ac{log} norm of the input image and $\tilde{P}$ is the \ac{log} norm of the currently rendered scene. This ensures that new Gaussians are placed only where additional detail is needed, rather than in already well-reconstructed areas. This direct sampling approach replaces the iterative densification process commonly used in 3D Gaussian Splatting with a more immediate placement strategy that positions Gaussians based on image content analysis.

To determine the depth at which sampled Gaussians should be placed and to provide depth supervision during optimization, we use Depth-Anything-2~\cite{yang2024depthv2} for monocular and Fast-ACVNet~\cite{xu2023accurateefficientstereomatching} for stereo depth estimation. Next, we follow the approach of Meuleman et al.~\cite{Meuleman_2025}, where the depth is then aligned to triangulated matches and refined through guided stereo matching to correct for monocular depth errors. 

\subsection{{Losses}}

We optimize Gaussian parameters through differentiable rendering using a multi-component loss:
\begin{equation}\label{equ:total_loss}
        \mathcal{L} = \mathcal{L}_{image} + \lambda_{depth}\mathcal{L}_{depth}
\end{equation} 

For the image loss, the L1 term measures pixel-wise differences, and the \ac{ssim} term captures structural similarities between rendered and ground truth images, balanced by $\lambda_s$.

\begin{equation}\label{equ:render_loss}
\mathcal{L}_{image} = (1-\lambda_{s})\left|{ I - I_{gt} }\right|_1 + \lambda_{s}{(1-\text{SSIM}(I, I_{gt}))}
\end{equation}

To enforce geometric accuracy, we implement a depth loss function that measures the absolute difference between the rendered and ground truth depth maps:
\begin{equation}\label{equ:depth_loss}
        \mathcal{L}_{depth} =  \left|{ D_\text{r} - D_\text{gt} }\right|_1
\end{equation} 

\subsection{{Chunk \ac{3dgs} Mapping}}

\subsubsection*{{Chunking}}\label{sec:chunking}
Our system employs a chunk-based scene management approach to efficiently render and optimize large-scale Gaussian splatting scenes. By partitioning the 3D space into discrete chunks, each containing a subset of Gaussians, we achieve selective loading, processing, and memory management that scales to massive environments while maintaining interactive performance.

The 3D world space is divided into regular cubic chunks of size $s$. We can determine into which chunk each Gaussian with position $p$ falls using centered chunks:
\begin{equation}\label{equ:grid_coord}
\operatorname{ChunkCoord}(p) = \left(\left\lfloor\frac{p_i + s/2}{s}\right\rfloor\right)_{i \in \{x,y,z\}}
\end{equation}

To enable $O(1)$ chunk location queries, chunk coordinates are encoded into a single 64-bit integer using a shifted base conversion. Each coordinate is allocated 21 bits after adding an offset of $2^{20}$ to handle negative values, providing a range of $\pm 1$M chunks per axis.

\begin{equation}
\text{EncodedID} = (c_x + 2^{20}) \cdot 2^{42} + (c_y + 2^{20}) \cdot 2^{21} + (c_z + 2^{20})
\end{equation}

\subsubsection*{Frustum Culling}
Our system employs hierarchical frustum culling with spatial subdivision to efficiently determine visible chunks for each keyframe. We extract frustum planes from the view-projection matrix and recursively test chunk regions at multiple levels, rejecting entire regions outside the frustum early and subdividing intersecting regions until reaching individual chunks. The process is parallelized using OpenMP and includes distance-based filtering and pose-aware caching to minimize redundant computations.

\begin{figure*}[htb]
    \centering
    \scriptsize
    \setlength{\tabcolsep}{1pt}
	\renewcommand{\arraystretch}{0.8}
	\newcommand{\sz}{0.31}
	\newcommand{\sh}{1.9cm}
    \begin{tabular}{ccccc}
        \rotatebox{90}{\hspace{12pt} RGB} &
        \includegraphics[width=\sz\linewidth]{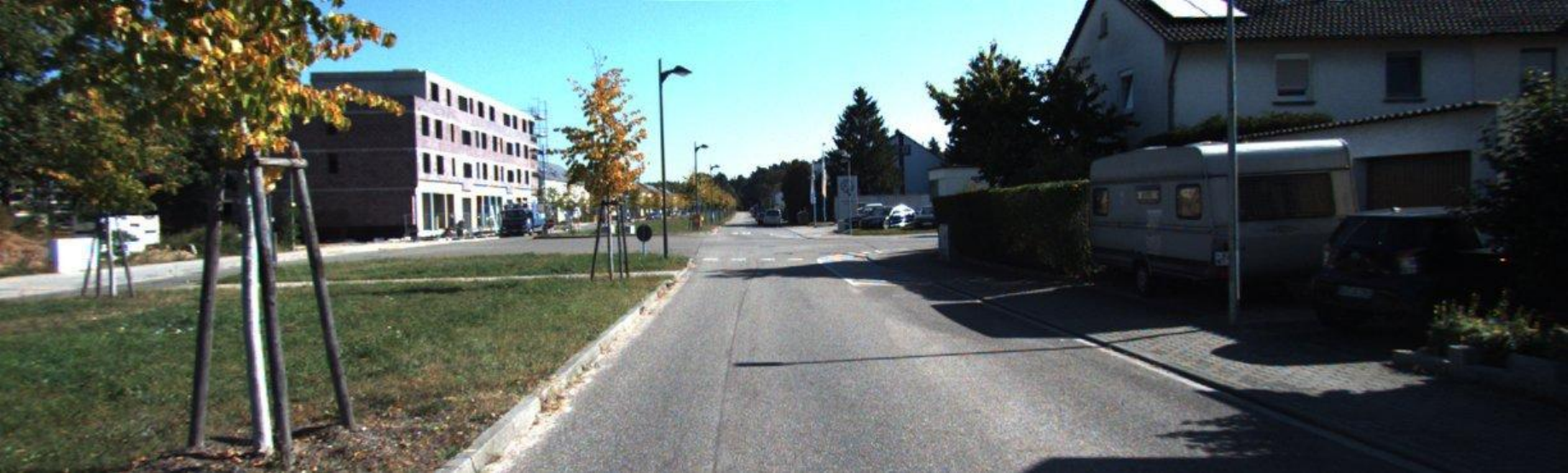}
        \includegraphics[width=\sz\linewidth]{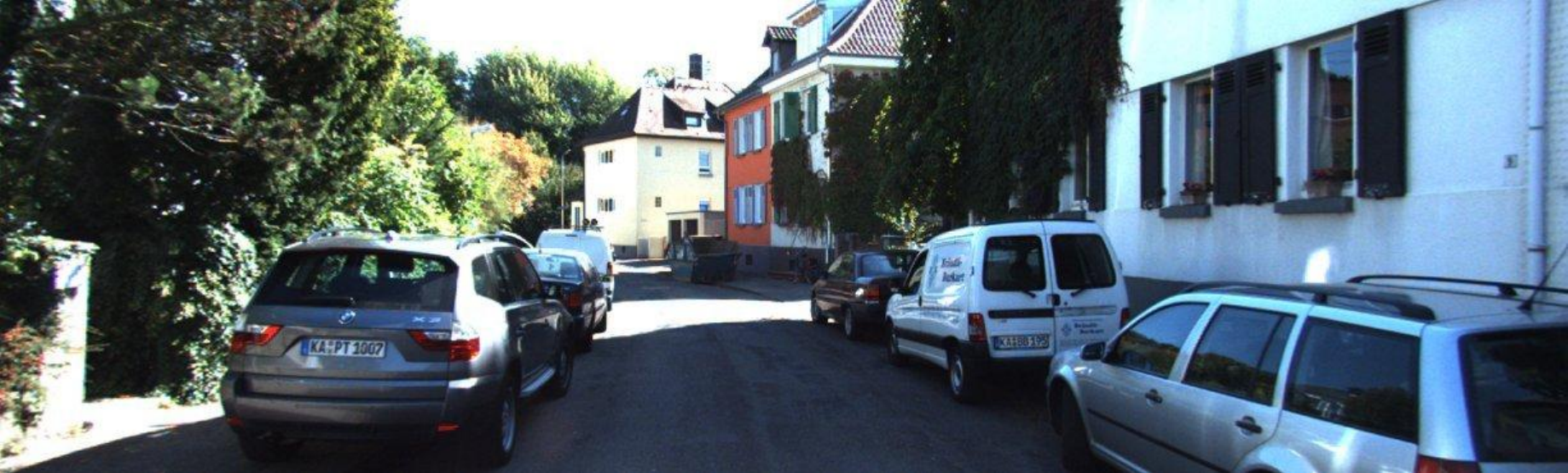}
        \includegraphics[width=\sz\linewidth]{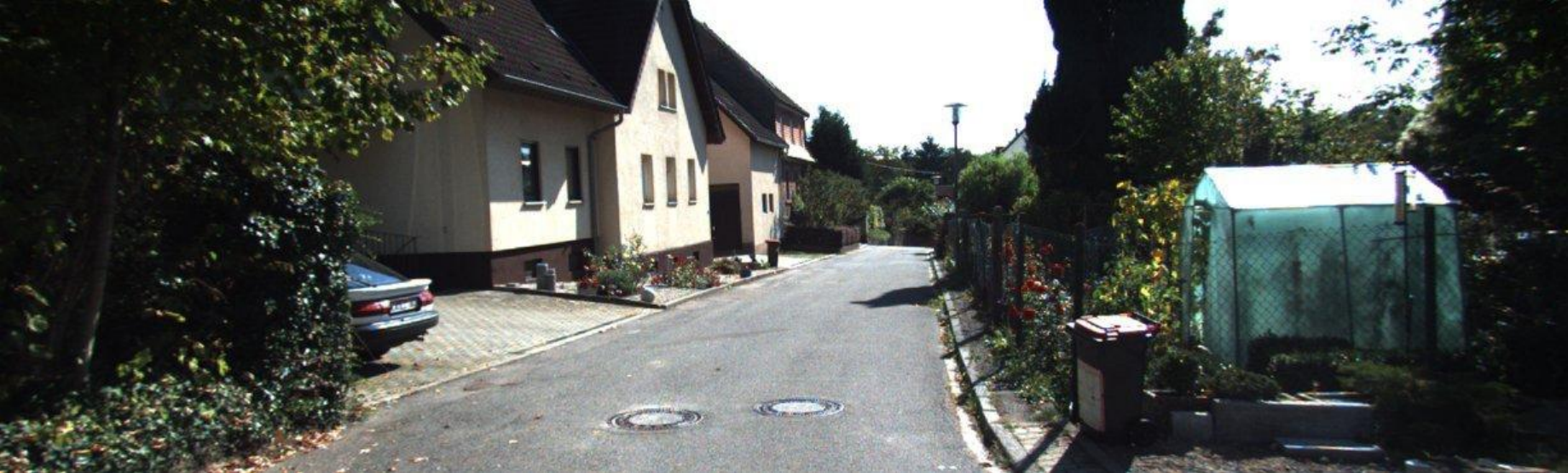}
        \\
        \rotatebox{90}{\hspace{11pt} CaRtGS} &
        \includegraphics[width=\sz\linewidth]{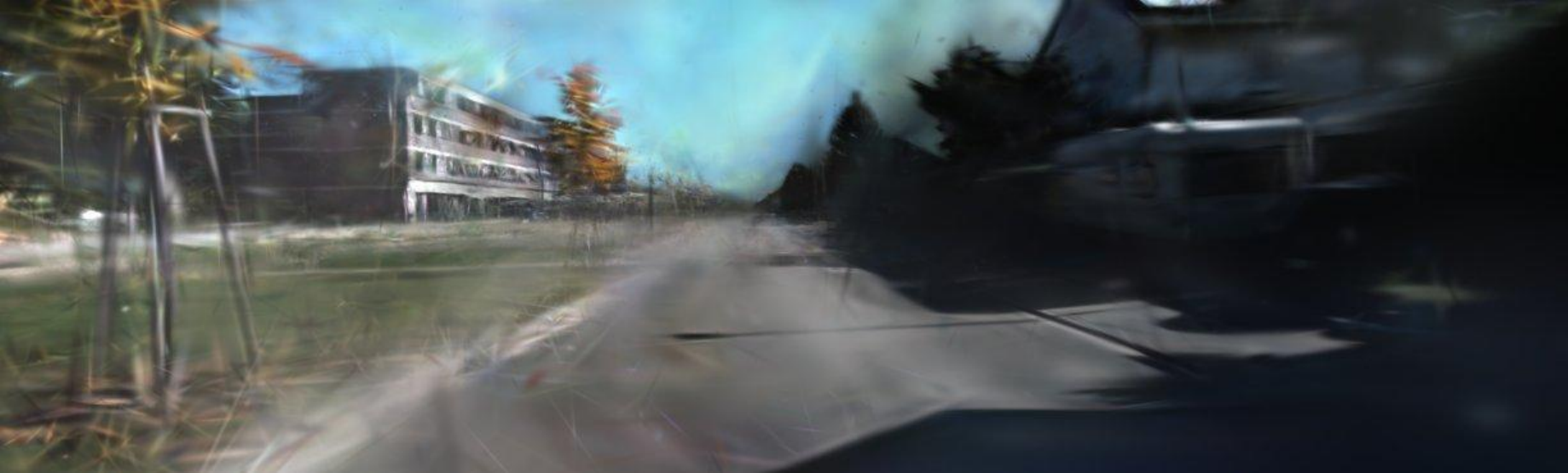}
        \includegraphics[width=\sz\linewidth]{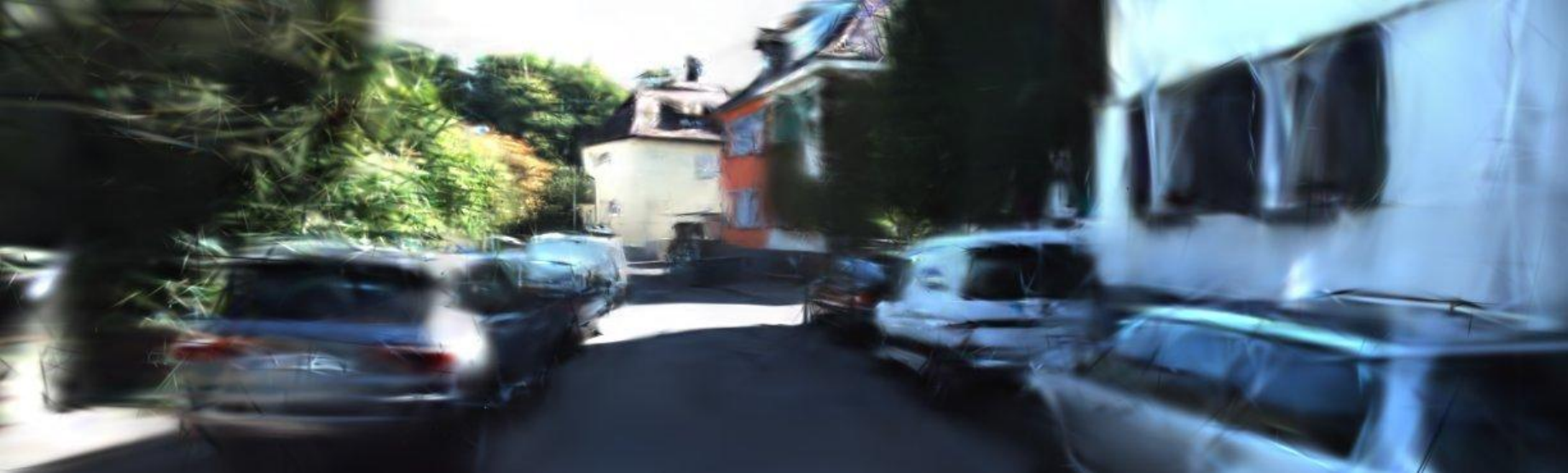}
        \includegraphics[width=\sz\linewidth]{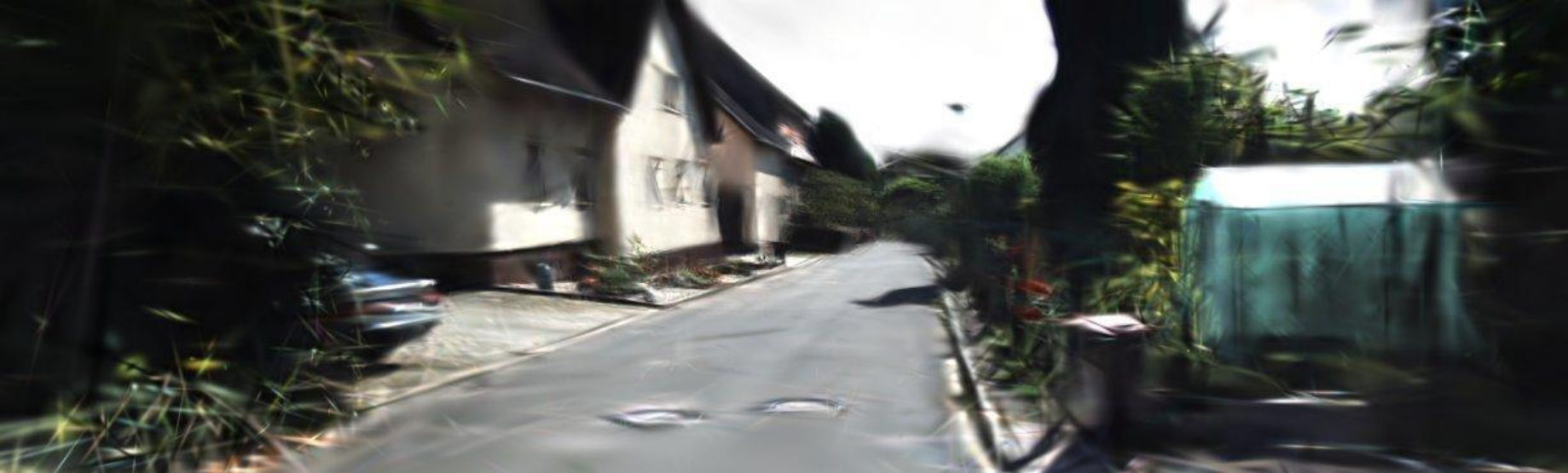}
        \\
        \rotatebox{90}{\hspace{6pt} On-The-Fly} &
        \includegraphics[width=\sz\linewidth]{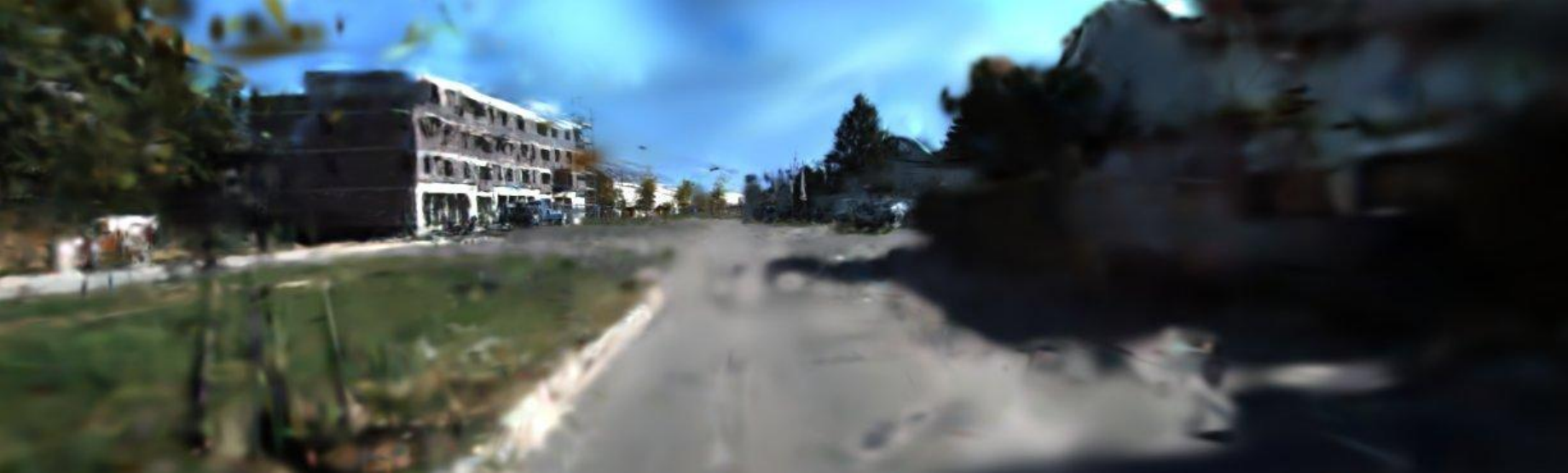}
        \includegraphics[width=\sz\linewidth]{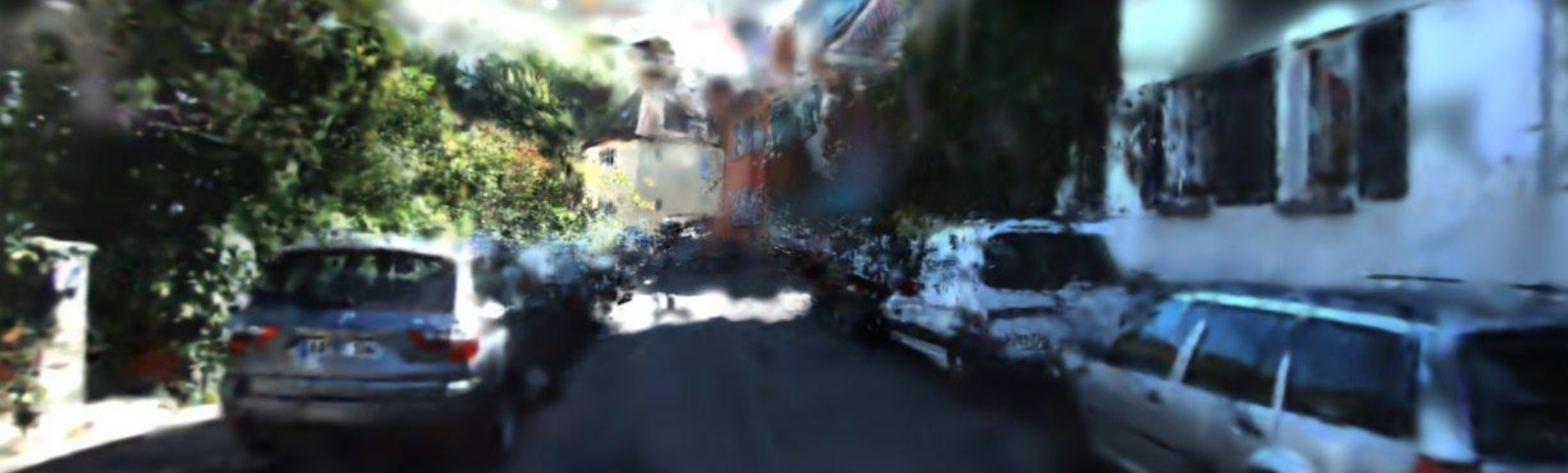}
        \includegraphics[width=\sz\linewidth]{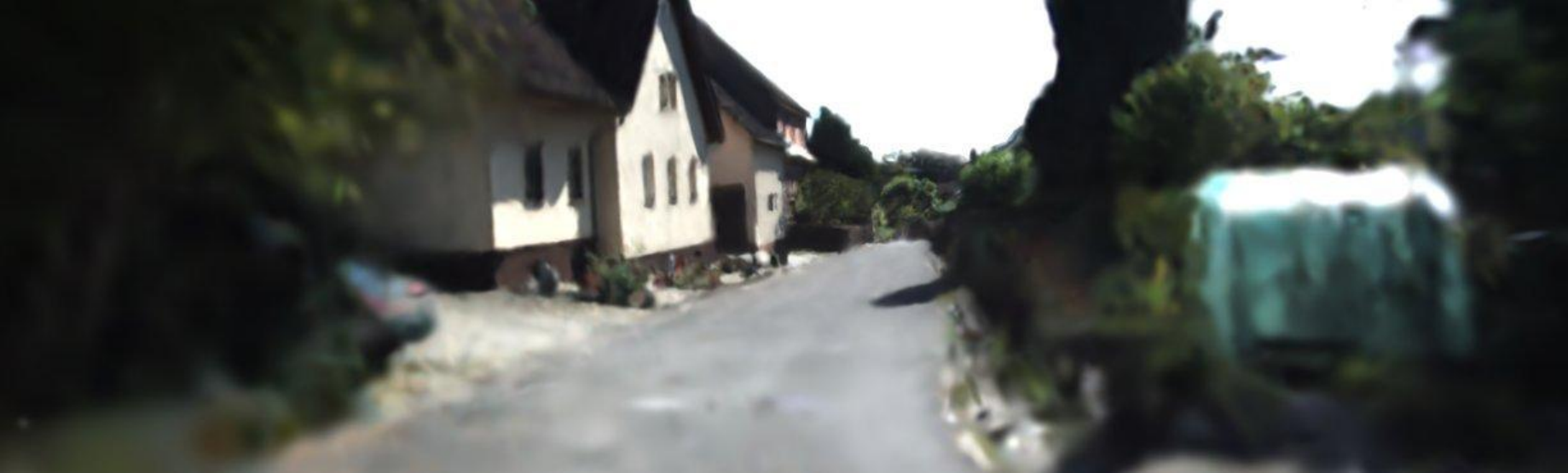}
        \\
        \rotatebox{90}{\hspace{5pt} GigaSLAM} &
        \includegraphics[width=\sz\linewidth]{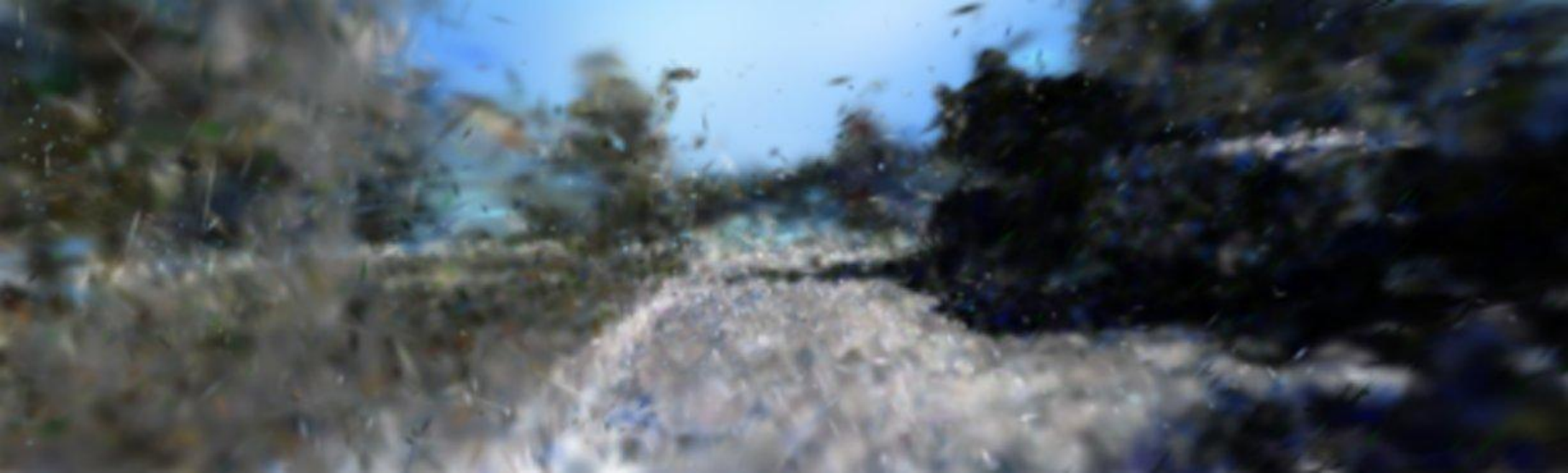}
        \includegraphics[width=\sz\linewidth]{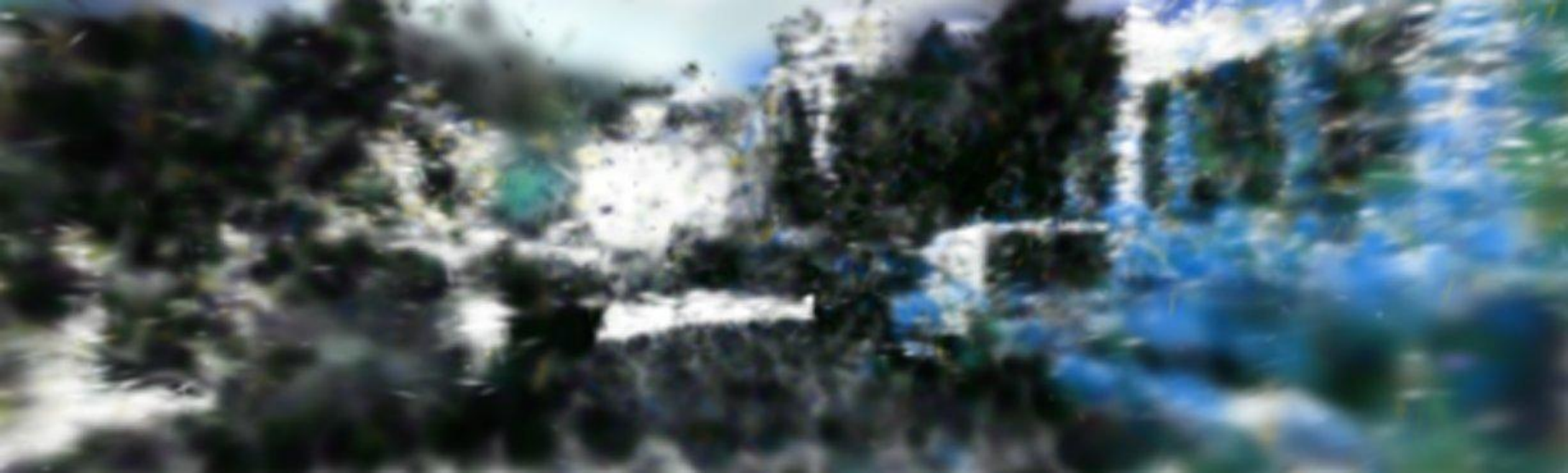}
        \includegraphics[width=\sz\linewidth]{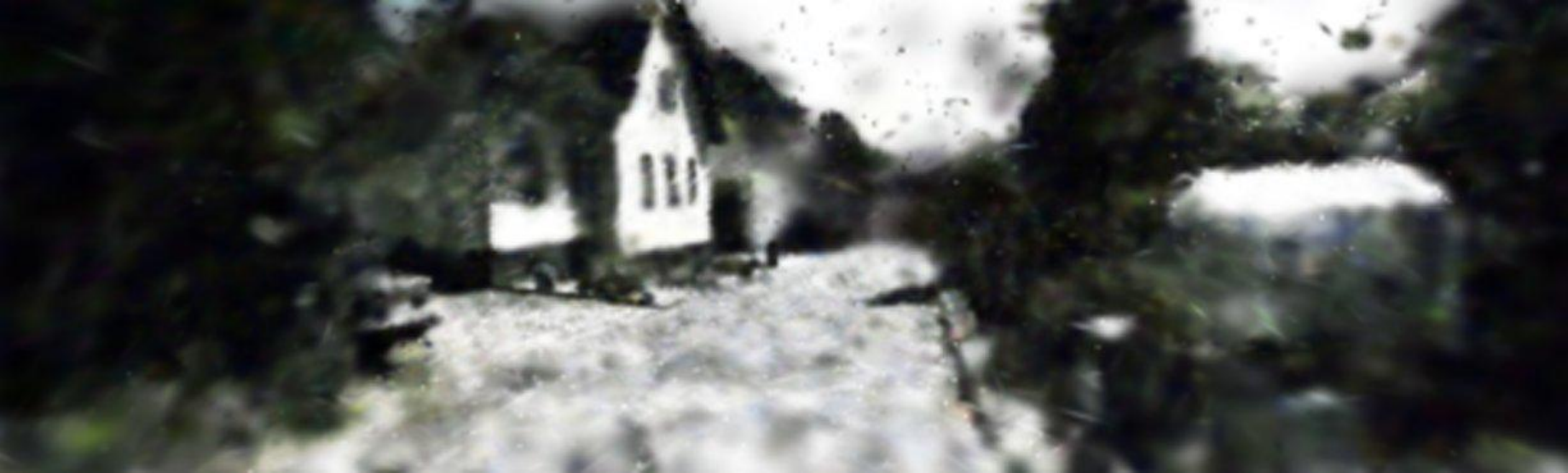}
        \\
         \rotatebox{90}{\hspace{13pt} \textbf{Ours}} &
        \includegraphics[width=\sz\linewidth]{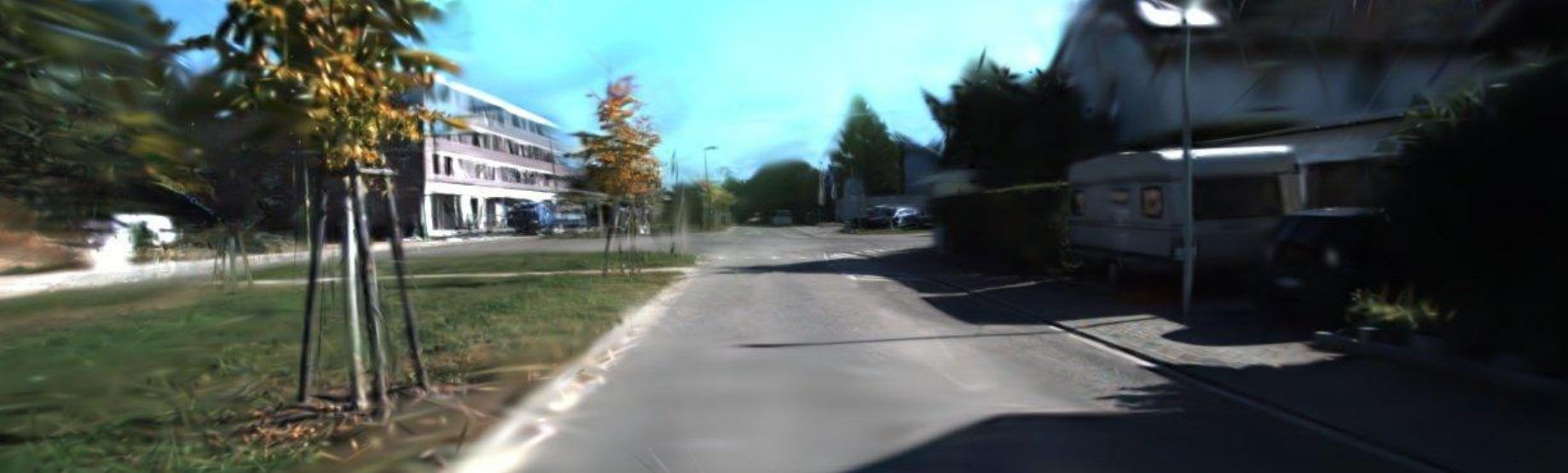}
        \includegraphics[width=\sz\linewidth]{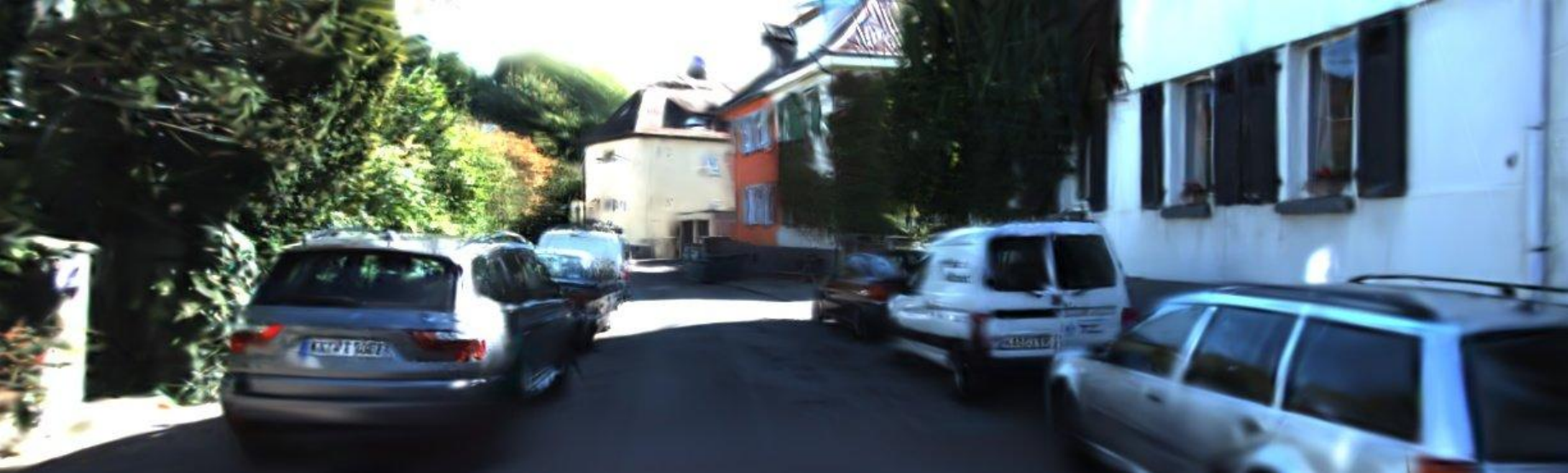}
        \includegraphics[width=\sz\linewidth]{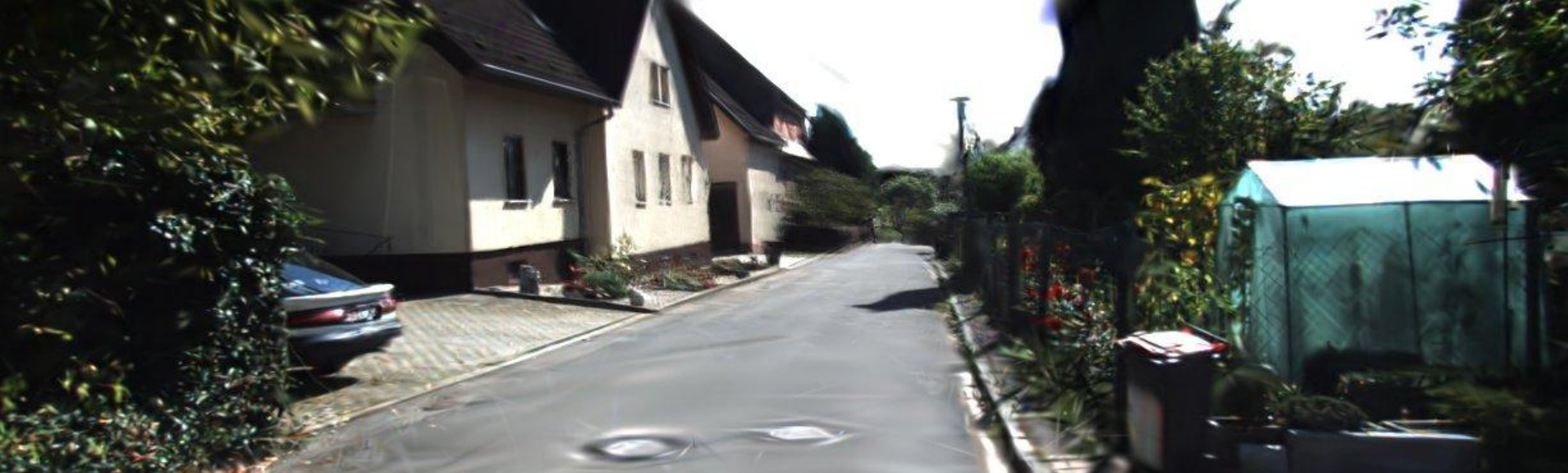}
        \\
        % Add scene labels at the bottom
        & 
        \makebox[\sz\linewidth]{\textbf{Scene 06}}\hspace{1pt}
        \makebox[\sz\linewidth]{\textbf{Scene 07}}\hspace{1pt}
        \makebox[\sz\linewidth]{\textbf{Scene 10}}
        \\
        
    \end{tabular}
    \caption{\textbf{Qualitative results on the KITTI~\cite{Geiger2012CVPR} dataset.} Reconstruction results on three scenes by all methods. CaRtGS exhibits floating artifacts, potentially caused by missing depth supervision. On-The-Fly suffers from tracking drift and lacks loop closure. GigaSLAM's neural approach shows incomplete convergence without its post-processing refinement. Our method achieves superior quality through robust tracking, depth-supervised Gaussian placement, and efficient chunk-based optimization.}
    \label{fig:qualitative_kitti}
\end{figure*}

\subsection{Optimization}
Our system follows a keyframe-driven architecture where each selected keyframe determines which chunks are loaded into VRAM based on its visible set. The optimization pass follows a structured workflow (visualized in Fig. \ref{fig:pipeline}). The process begins by selecting a keyframe and performing frustum culling to identify visible chunks. These visible chunks are then loaded into \ac{vram} if not already present. 

To maintain memory efficiency, our system operates under a configurable \ac{vram} budget, which we set to 1.5 million Gaussians for our experiments. When loading new chunks would cause the system to exceed this limit, we evict and save existing chunks from \ac{vram} to disk using a \ac{lru} principle. After ensuring all necessary chunks are in memory, we use the chunk visibility mask to index only the visible Gaussians for rendering. These selected Gaussians are passed to the rasterizer, which renders RGB and depth images from the selected keyframe. The rendered images are then used to compute gradients and optimize the Gaussian parameters according to the loss function defined in Equation~\ref{equ:total_loss}.

\subsubsection*{\ac{lru} Eviction Strategy}

The \ac{lru} eviction operates as greedy selection. Given evictable chunks sorted by access time (oldest first), we select the minimum number of chunks needed to accommodate incoming data. This ensures minimal disruption by preserving recently accessed regions and maintaining spatial coherence.

In addition to chunk eviction, we maintain an independent LRU queue with a 400-keyframe budget. Keyframes are loaded on demand; when the budget is exceeded, the least recently used are saved to disk and evicted.

\subsection{Keyframe Selection}
Intelligent keyframe selection is paramount for our chunk-based \ac{3dgs} architecture. The fundamental challenge lies in balancing two competing objectives: maintaining spatial locality for efficient I/O operations while ensuring sufficient gradient diversity for stable optimization. Poor keyframe selection can lead to computational thrashing due to excessive chunk swapping or catastrophic forgetting due to insufficient viewing constraints.

Let $\mathcal{K}_t$ denote the set of available keyframes at time $t$, where each keyframe $k \in \mathcal{K}_t$ has an associated 3D position $p_k$. Let $\mathcal{C}_{active}$ represent the set of chunks currently loaded in \ac{vram}, and $\mathcal{C}_v(k)$ denote the set of chunks visible from keyframe $k$. When operating near our \ac{vram} budget limit of 1.5 million Gaussians, the chunk overlap between a candidate keyframe's visible chunks and the currently active set becomes critical for I/O efficiency:
\begin{equation}
\text{Overlap}(k) = \frac{|\mathcal{C}_v(k) \cap \mathcal{C}_{active}|}{|\mathcal{C}_v(k)|}
\end{equation}

This metric measures the fraction of chunks required by keyframe $k$ that are already resident in \ac{vram}. High overlap values (approaching 1.0) indicate minimal loading requirements, while low overlap values necessitate loading many new chunks. In large-scale scenes operating near the VRAM budget, low overlap triggers our \ac{lru} eviction mechanism, causing excessive chunk swapping that can severely degrade training throughput due to I/O bottlenecks.

To address the fundamental tension between spatial locality and gradient diversity, we organize keyframes into a spatial grid with resolution $g = 200$ m (Figure~\ref{fig:pipeline}, top-right). This keyframe selection grid operates independently from and at a coarser resolution than the chunk grid used for Gaussian partitioning. For any query position $p$, we define the spatial candidate set as:
\begin{equation}
\mathcal{K}(p) = \{k \in \mathcal{K}_t : \lfloor p_k/g \rfloor = \lfloor p/g \rfloor\}
\end{equation}
where $\lfloor \cdot/g \rfloor$ applies element-wise division and floor to discretize 3D positions into grid coordinates at resolution $g$. This spatial hashing ensures that all keyframes in $\mathcal{K}(p)$ lie within the same grid cell, promoting spatial locality.

At each optimization step, we identify the spatial candidate set $\mathcal{K}(p_{latest})$ containing all keyframes within the same grid cell as the most recently added keyframe at position $p_{latest}$. Within this spatially constrained set, we apply the usage-based, loss-weighted selection strategy from CaRtGS~\cite{Feng_2025}, which prioritizes keyframes with remaining usage allocations and allocates additional usage to high-loss keyframes to adaptively focus on challenging views.

This spatial grid approach provides several key advantages: when revisiting previously mapped regions, all historically relevant keyframes within the spatial locality contribute their viewing constraints regardless of temporal distance, preventing catastrophic forgetting. By constraining selection to keyframes within spatial grid cell $g$, we minimize I/O operations through high chunk overlap.

\subsection{Loop Closure}
While traditional voxel-based systems often require expensive global optimization or complete rebuilding of affected regions during loop closure, our chunked \ac{3dgs} system enables more targeted corrections. Our approach identifies affected keyframes and selectively transforms only the visible Gaussian points in relevant chunks.

\subsubsection*{Cross-Chunk Transformation}
For each affected keyframe, we perform frustum culling to identify visible chunks and apply a pose-correction transformation to both the positions and rotations of their Gaussians.

To minimize I/O overhead, we collect unique chunks across affected keyframes and estimate the total number of Gaussians requiring transformation. If this remains within our memory budget, we batch-load all chunks in a single operation and use a global transformation mask to prevent redundant processing of Gaussians visible across multiple keyframes. For extensive loop closures exceeding memory constraints, we process keyframes sequentially using per-keyframe chunk loading.

\subsubsection*{Chunk Redistribution}
After transformation, Gaussians may cross chunk boundaries. We handle this by recomputing chunk assignments based on updated positions and redistributing moved Gaussians to their correct chunks, ensuring the spatial hierarchy remains consistent.

\subsubsection*{Post-Correction Refinement}
After geometric transformation and redistribution, we perform targeted optimization to refine the affected regions. We pause new frame ingestion, and for keyframes where independent reconstructions merge (loop-closure detection points), we reset the Gaussian opacity and optimizer states to enable fresh optimization of the junction region. This extended refinement runs for 1 k iterations (approximately 10 seconds), ensuring improved blending between previously disconnected map sections.

\begin{table}[tb]
\caption{\textbf{SLAM Tracking Accuracy on KITTI~\cite{Geiger2012CVPR} dataset (ATE in meters)} $\downarrow$.
\fscap{Best} and \ndcap{second best} results are highlighted. \\
\textcolor{red}{$\times$} = OOM crash, \textbf{--} = tracking loss. 
S = Stereo, M = Mono. \\ 
OTF = On-The-Fly~\cite{Meuleman_2025}, Giga = GigaSLAM~\cite{deng2025gigaslamlargescalemonocularslam}}
\setlength{\tabcolsep}{3pt} 
\renewcommand{\arraystretch}{1.0}
\begin{centering}
\begin{tabular}{lccccc}
\hline 
\textbf{Seq.} & \textbf{CaRtGS (S)} & \textbf{OTF (M)} & \textbf{Giga (M)} & \textbf{Ours-5 (S)} & \textbf{Ours-20 (S)}\tabularnewline
\hline 
% 0
\textbf{00} & \textcolor{red}{$\times$} & 20.90 & \textcolor{red}{$\times$} & \fs{0.82} & \nd{0.89} \tabularnewline
% 1
\textbf{01} & 26.14 & \textbf{--} & 74.48 & \nd{16.41} & \fs{9.99} \tabularnewline
% 2
\textbf{02} & \textcolor{red}{$\times$} & \textbf{--} & \textcolor{red}{$\times$} & \nd{4.50} & \fs{4.05} \tabularnewline
% 3
\textbf{03} & 0.37 & \textbf{--} & 1.49 & \fs{0.32} & \nd{0.33} \tabularnewline
% 4
\textbf{04} & \fs{0.17} & \textbf{--} & \nd{1.78} & \fs{0.17} & \fs{0.17} \tabularnewline
% 5
\textbf{05} & 0.43 & 2.38 & \textcolor{red}{$\times$} & \nd{0.41} & \fs{0.37} \tabularnewline
% 6
\textbf{06} & \fs{0.42} & 9.69 & 1.20 & \nd{0.58} & 0.72 \tabularnewline
% 7
\textbf{07} & 0.44 & 21.21 & 3.41 & \fs{0.35} & \nd{0.40} \tabularnewline
% 8
\textbf{08} & \textcolor{red}{$\times$} & \textbf{--} & \textcolor{red}{$\times$} & \fs{2.99} & \nd{3.18} \tabularnewline
% 9
\textbf{09} & \fs{0.99} & \textbf{--} & 3.86 & \fs{0.99} & \nd{1.02} \tabularnewline
% 10
\textbf{10} & \nd{1.34} & 19.75 & 2.45 & \nd{1.34} & \fs{1.18} \tabularnewline

\hline 
\end{tabular}
\par\end{centering}
\label{tab:kitti_tracking_ate}
\end{table}

\begin{table}[tb]
\caption{\textbf{Processing Speed on KITTI~\cite{Geiger2012CVPR} dataset (Processing FPS)} $\uparrow$.
\fscap{Best} and \ndcap{second best} results are highlighted. \\ 
\textcolor{red}{$\times$} = OOM crash, \textbf{--} = tracking loss. 
S = Stereo, M = Mono. \\ 
OTF = On-The-Fly~\cite{Meuleman_2025}, Giga = GigaSLAM~\cite{deng2025gigaslamlargescalemonocularslam}}
\setlength{\tabcolsep}{3pt} 
\renewcommand{\arraystretch}{1.0}
\begin{centering}
\begin{tabular}{lccccc}
\hline 
\textbf{Seq.} & \textbf{CaRtGS (S)} & \textbf{OTF (M)} & \textbf{Giga (M)} & \textbf{Ours-5 (S)} & \textbf{Ours-20 (S)}\tabularnewline
\hline 
\textbf{00} & \textcolor{red}{$\times$} & \fs{11.39} & \textcolor{red}{$\times$} & 1.51 & \nd{4.94} \tabularnewline
\textbf{01} & 0.59 & \textbf{--} & \fs{3.30} & 0.59 & \nd{2.26} \tabularnewline
\textbf{02} & \textcolor{red}{$\times$} & \textbf{--} & \textcolor{red}{$\times$} & \nd{1.15} & \fs{3.69} \tabularnewline
\textbf{03} & 1.80 & \textbf{--} & \nd{2.55} & 1.81 & \fs{6.54} \tabularnewline
\textbf{04} & 0.88 & \textbf{--} & \nd{2.25} & 0.89 & \fs{3.35} \tabularnewline
\textbf{05} & 1.59 & \fs{15.85} & \textcolor{red}{$\times$} & 1.53 & \nd{4.69} \tabularnewline
\textbf{06} & 1.15 & \fs{11.02} & 1.88 & 1.13 & \nd{3.95} \tabularnewline
\textbf{07} & 2.11 & \fs{11.83} & 2.26 & 2.05 & \nd{6.59} \tabularnewline
\textbf{08} & \textcolor{red}{$\times$} & \textbf{--} & \textcolor{red}{$\times$} & \nd{1.61} & \fs{5.82} \tabularnewline
\textbf{09} & 1.20 & \textbf{--} & \nd{1.85} & 1.18 & \fs{4.16} \tabularnewline
\textbf{10} & 1.65 & \fs{12.17} & 2.09 & 1.66 & \nd{5.90} \tabularnewline
\hline 
\end{tabular}
\par\end{centering}
\label{tab:kitti_tracking_fps}
\end{table}

\begin{table*}[tb]
\caption{\textbf{Rendering Quality on KITTI~\cite{Geiger2012CVPR}.} Novel view synthesis comparison. \textbf{--} indicates tracking loss and \textcolor{red}{$\times$} indicates out-of-memory. \fscap{Best} and \ndcap{second best} results are highlighted.}
\setlength{\tabcolsep}{7pt} 
\begin{centering}
\begin{tabular}{ccccccccccccc}
\hline 
\textbf{Method} & \textbf{Metric} & \textbf{00} & \textbf{01} & \textbf{02} & \textbf{03} & \textbf{04} & \textbf{05} & \textbf{06} & \textbf{07} & \textbf{08} & \textbf{09} & \textbf{10}\tabularnewline
Frames & - & 4541 & 1101 & 4661 & 801 & 271 & 2761 & 1101 & 1101 & 4071 & 1591 & 1201\tabularnewline
Length (km) & - & 3.72 & 2.45 & 5.07 & 0.56 & 0.39 & 2.21 & 1.23 & 0.65 & 3.22 & 1.71 & 0.92\tabularnewline
Loop Closure & - & \checkmark & $\times$ & \checkmark & $\times$ & $\times$ & \checkmark & \checkmark & \checkmark & \checkmark & \checkmark & $\times$\tabularnewline
\hline 
%-----------------------------------------------------
\multirow{3}{*}{%
\begin{tabular}{c}
{CaRtGS~\cite{Feng_2025}}\tabularnewline
{(Stereo)}\tabularnewline
\end{tabular}} & PSNR $\uparrow$ & \textcolor{red}{$\times$} & \fs{23.14} & \textcolor{red}{$\times$} & \nd{19.81} & \nd{21.97} & {18.16} & {17.92} & {17.43} & \textcolor{red}{$\times$} & 18.29 & {19.97} \tabularnewline
& SSIM $\uparrow$ & \textcolor{red}{$\times$} & \nd{0.74} & \textcolor{red}{$\times$} & {0.56} & {0.72} & {0.58} & 0.56 & 0.58 & \textcolor{red}{$\times$} & {0.56} & 0.63 \tabularnewline
& LPIPS $\downarrow$ & \textcolor{red}{$\times$} & \nd{0.31} & \textcolor{red}{$\times$} & 0.43 & {0.29} & 0.39 & 0.39 & 0.37 & \textcolor{red}{$\times$} & 0.43 & 0.37 \tabularnewline
\cline{1-13}
%-----------------------------------------------------
\multirow{3}{*}{%
\begin{tabular}{c}
{On-the-fly~\cite{Meuleman_2025}} \tabularnewline
{(Mono)}\tabularnewline
\end{tabular}}
& PSNR $\uparrow$ & {17.01} & \textbf{--} & \textbf{--} & \textbf{--} & \textbf{--} & 17.10 & {18.44} & {18.56} & \textbf{--} & \textbf{--} & 18.22 \tabularnewline
& SSIM $\uparrow$ & {0.59} & \textbf{--} & \textbf{--} & \textbf{--} & \textbf{--} & {0.58} & \nd{0.61} & {0.64} & \textbf{--} & \textbf{--} & 0.58 \tabularnewline
& LPIPS $\downarrow$ & 0.46 & \textbf{--} & \textbf{--} & \textbf{--} & \textbf{--} & 0.44 & 0.40 & {0.38} & \textbf{--} & \textbf{--} & 0.47 \tabularnewline
\cline{1-13}
%-----------------------------------------------------
\multirow{3}{*}{%
\begin{tabular}{c}
{GigaSLAM~\cite{deng2025gigaslamlargescalemonocularslam}}\tabularnewline
{(Mono)}\tabularnewline
\end{tabular}} & PSNR $\uparrow$ & \textcolor{red}{$\times$} & 15.86 & \textcolor{red}{$\times$} & 16.09 & 14.74 & \textcolor{red}{$\times$} & 14.72 & 14.53 & \textcolor{red}{$\times$} & 15.56 & 15.75 \tabularnewline
& SSIM $\uparrow$ & \textcolor{red}{$\times$} & 0.50 & \textcolor{red}{$\times$} & 0.39 & 0.33 & \textcolor{red}{$\times$} & 0.44 & 0.42 & \textcolor{red}{$\times$} & 0.41 & 0.44 \tabularnewline
& LPIPS $\downarrow$ & \textcolor{red}{$\times$} & 0.62 & \textcolor{red}{$\times$} & 0.64 & 0.63 & \textcolor{red}{$\times$} & 0.70 & 0.68 & \textcolor{red}{$\times$} & 0.66 & 0.68 \tabularnewline
\cline{1-13}
%-----------------------------------------------------
\multirow{3}{*}{%
\begin{tabular}{c}
\textbf{Ours 5 km/h}\tabularnewline
\textbf{(Stereo)}\tabularnewline
\end{tabular}} & PSNR $\uparrow$ & \fs{20.33} & \nd{22.07} & \fs{20.35} & \fs{20.84} & \fs{21.98} & \fs{20.08} & \nd{19.67} & \fs{20.09} & \fs{20.33} & \fs{21.15} & \fs{22.89} \tabularnewline
& SSIM $\uparrow$ & \fs{0.72} & \fs{0.77} & \fs{0.69} & \fs{0.68} & \fs{0.78} & \fs{0.70} & \fs{0.68} & \fs{0.74} & \fs{0.72} & \fs{0.73} & \fs{0.78} \tabularnewline
& LPIPS $\downarrow$ & \fs{0.26} & \fs{0.26} & \fs{0.28} & \fs{0.31} & \fs{0.23} & \fs{0.28} & \fs{0.31} & \fs{0.25} & \fs{0.27} & \fs{0.25} & \fs{0.22} \tabularnewline
\cline{1-13}
%-----------------------------------------------------
\multirow{3}{*}{%
\begin{tabular}{c}
\textbf{Ours 20 km/h}\tabularnewline
\textbf{(Stereo)}\tabularnewline
\end{tabular}} & PSNR $\uparrow$ & \nd{19.65} & 21.54 & \nd{19.67} & 19.15 & 20.76 & \nd{19.01} & \fs{19.75} & \nd{19.83} & \nd{19.30} & \nd{20.41} & \nd{21.63} \tabularnewline
& SSIM $\uparrow$ & \nd{0.70} & \fs{0.77} & \nd{0.67} & \nd{0.64} & \nd{0.75} & \nd{0.67} & \fs{0.68} & \nd{0.73} & \nd{0.69} & \nd{0.72} & \nd{0.75} \tabularnewline
& LPIPS $\downarrow$ & \nd{0.31} & \fs{0.26} & \nd{0.31} & \nd{0.35} & \nd{0.26} & \nd{0.31} & \nd{0.32} & \nd{0.26} & \nd{0.31} & \nd{0.28} & \nd{0.26} \tabularnewline
\hline 
\end{tabular}
\par\end{centering}
\label{tab:kitti_rendering}
\end{table*} 
\section{Experiments}

\begin{figure}[b]
        \hspace{-0.4cm}
    \centering
    \scriptsize
    \setlength{\tabcolsep}{1pt}
	\renewcommand{\arraystretch}{0.8}
	\newcommand{\sz}{0.31}
	\newcommand{\sh}{1.65cm}
    \begin{tabular}{cccc}
        \rotatebox{90}{\hspace{12pt} RGB} &
        \includegraphics[width=\sz\linewidth]{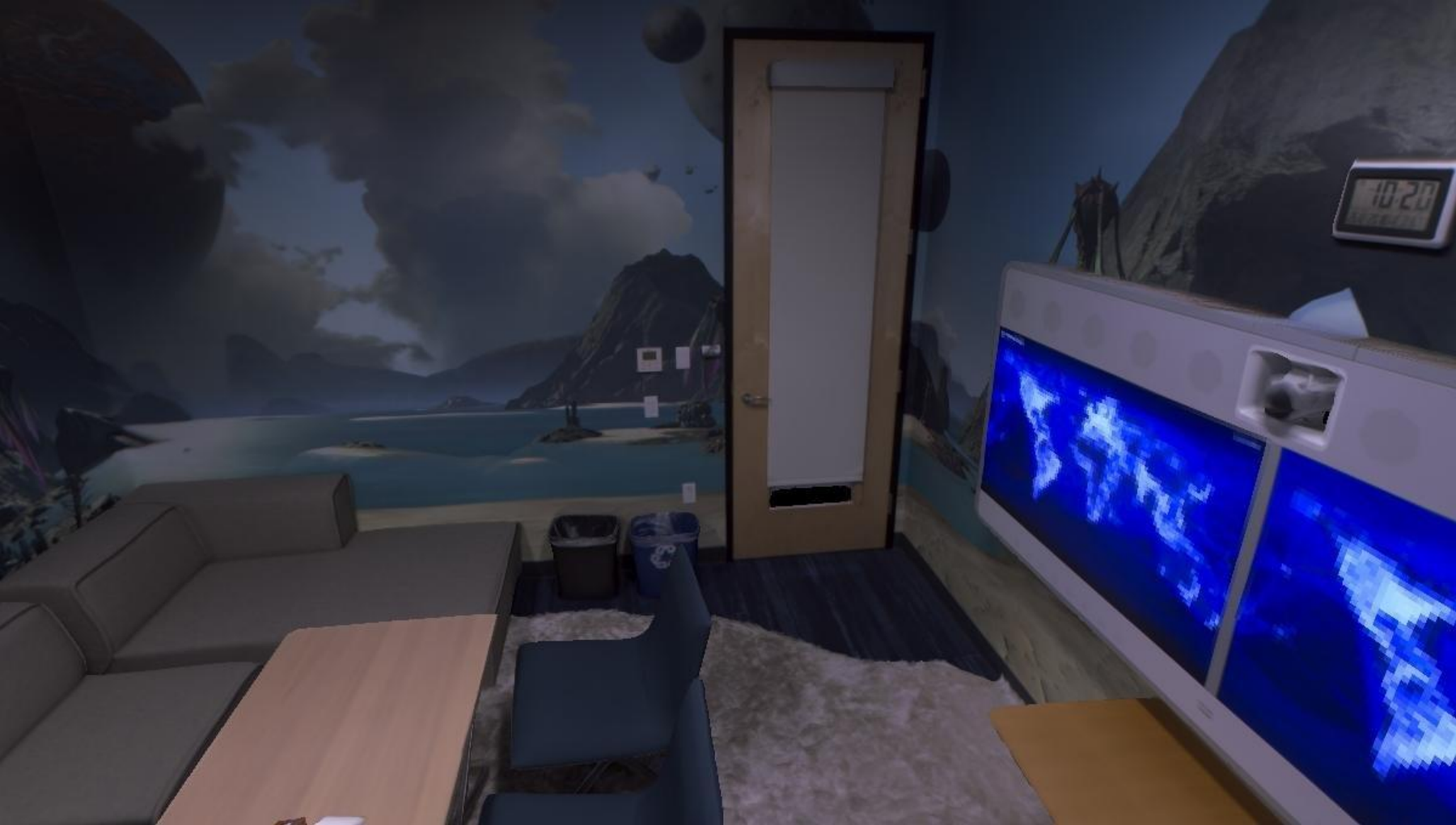} &
        \includegraphics[width=\sz\linewidth]{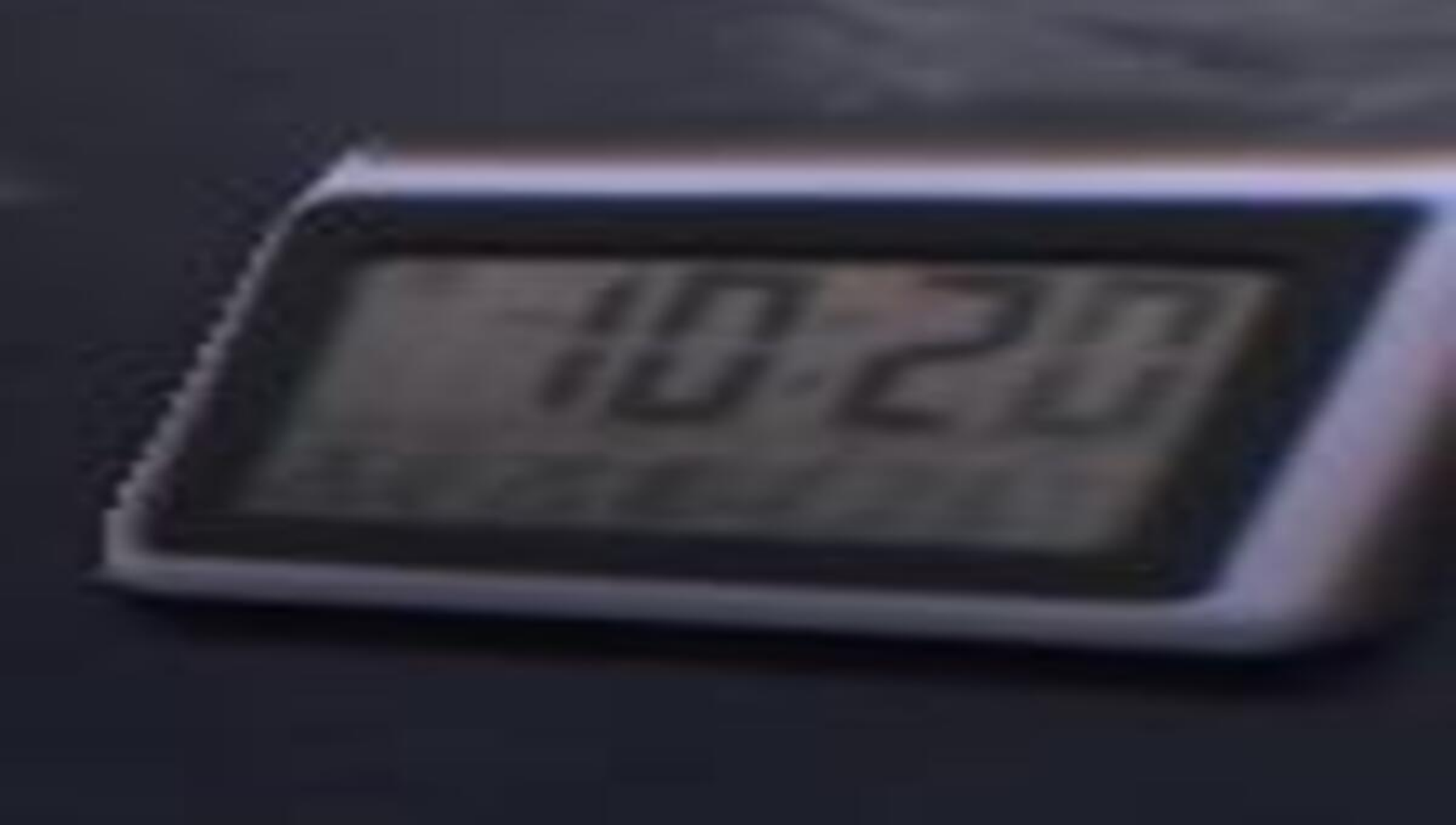} &
        \includegraphics[width=\sz\linewidth]{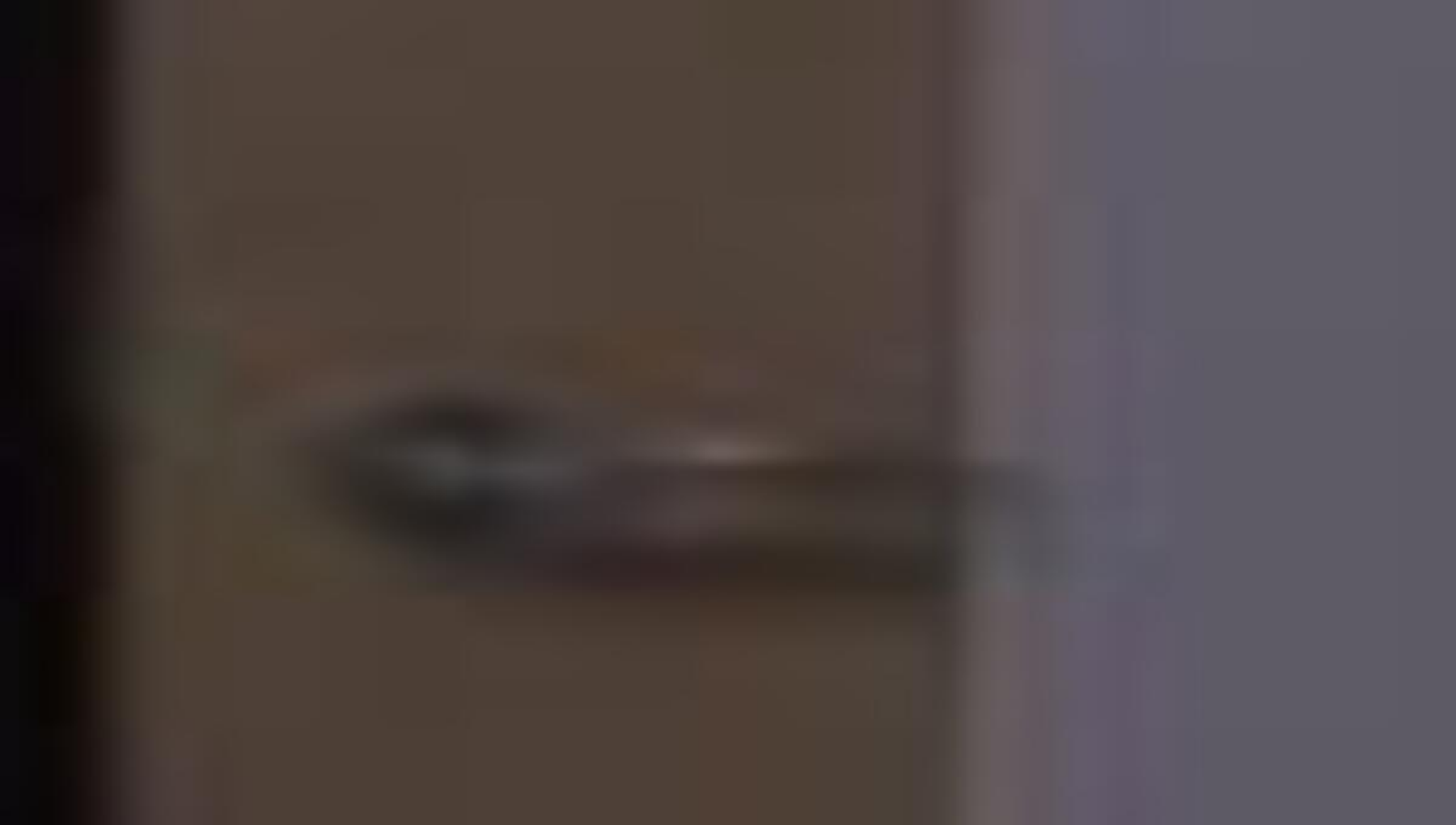}\\

        \rotatebox{90}{\hspace{8pt} CaRtGS} &
        \includegraphics[width=\sz\linewidth]{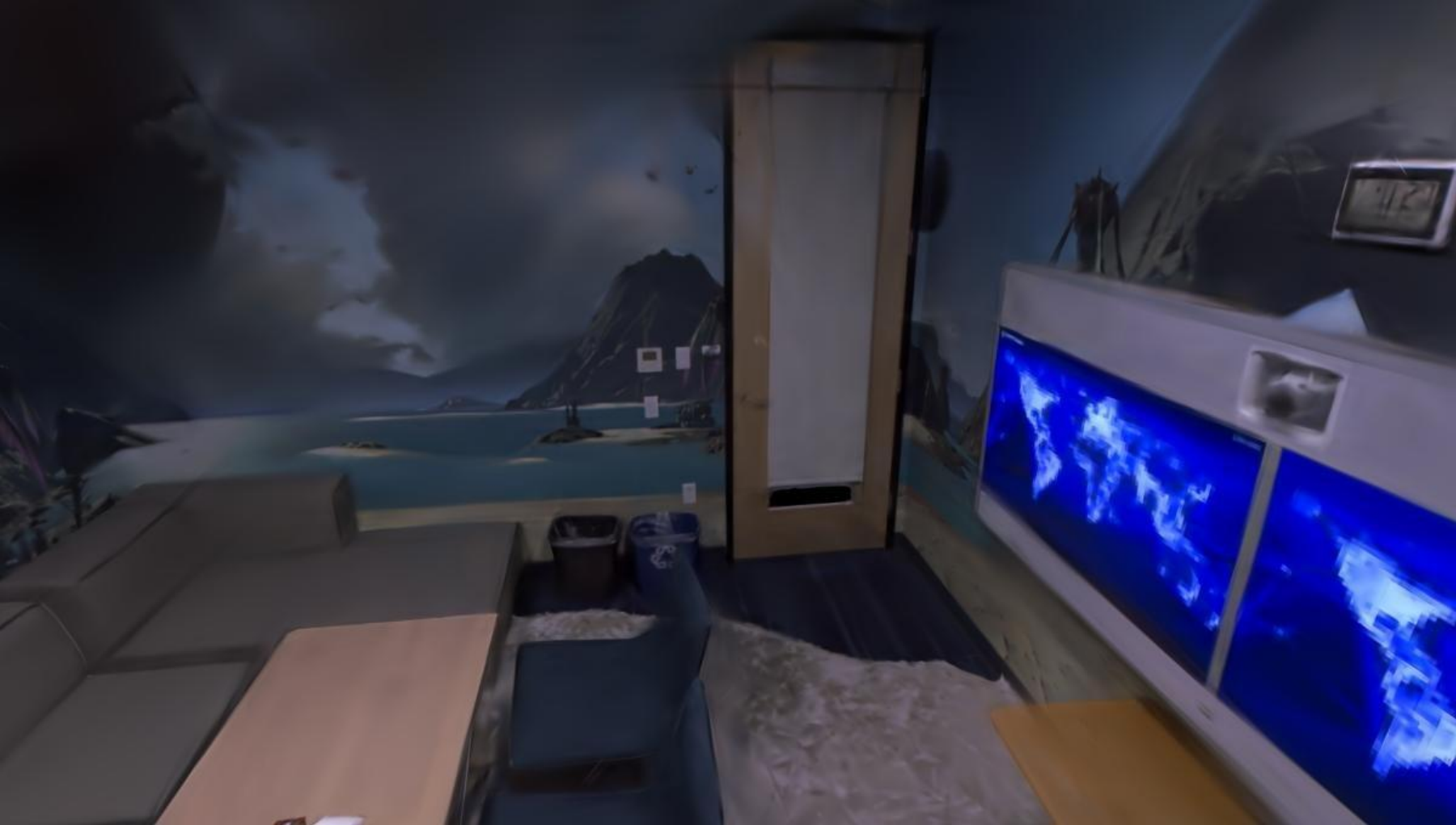} &
        \includegraphics[width=\sz\linewidth]{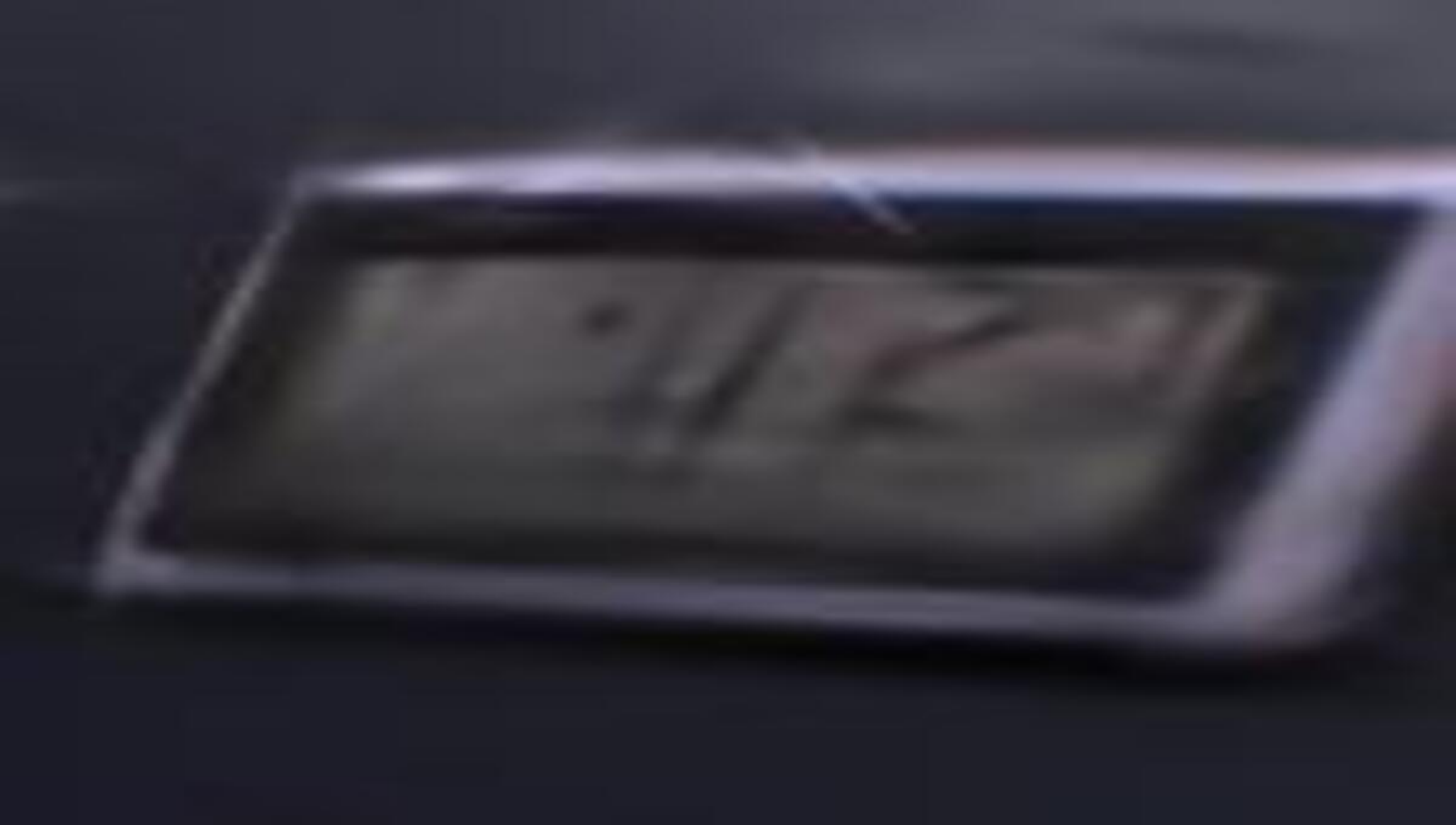} &
        \includegraphics[width=\sz\linewidth]{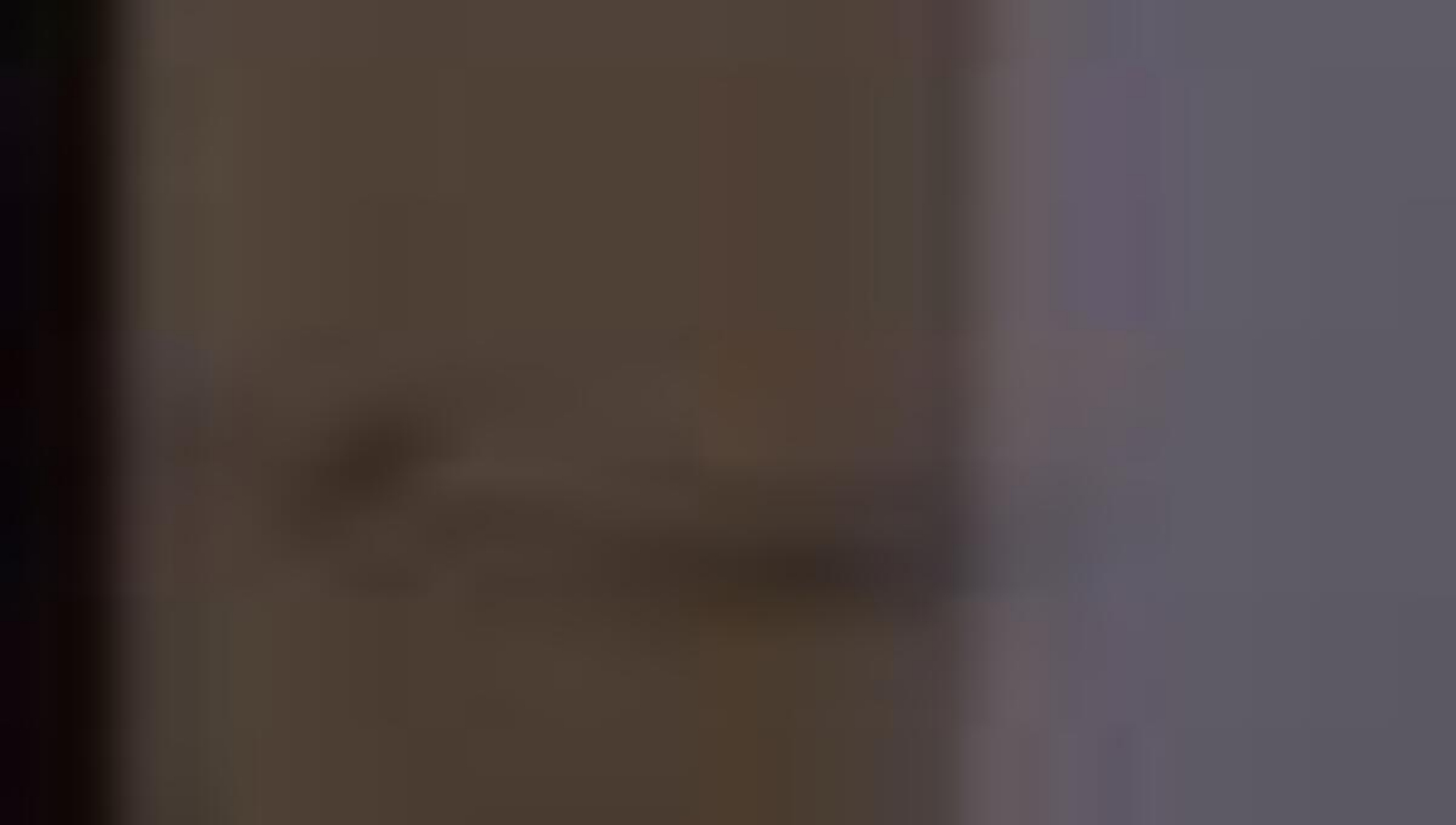}\\

        \rotatebox{90}{\hspace{2pt} On-The-Fly} &
        \includegraphics[width=\sz\linewidth]{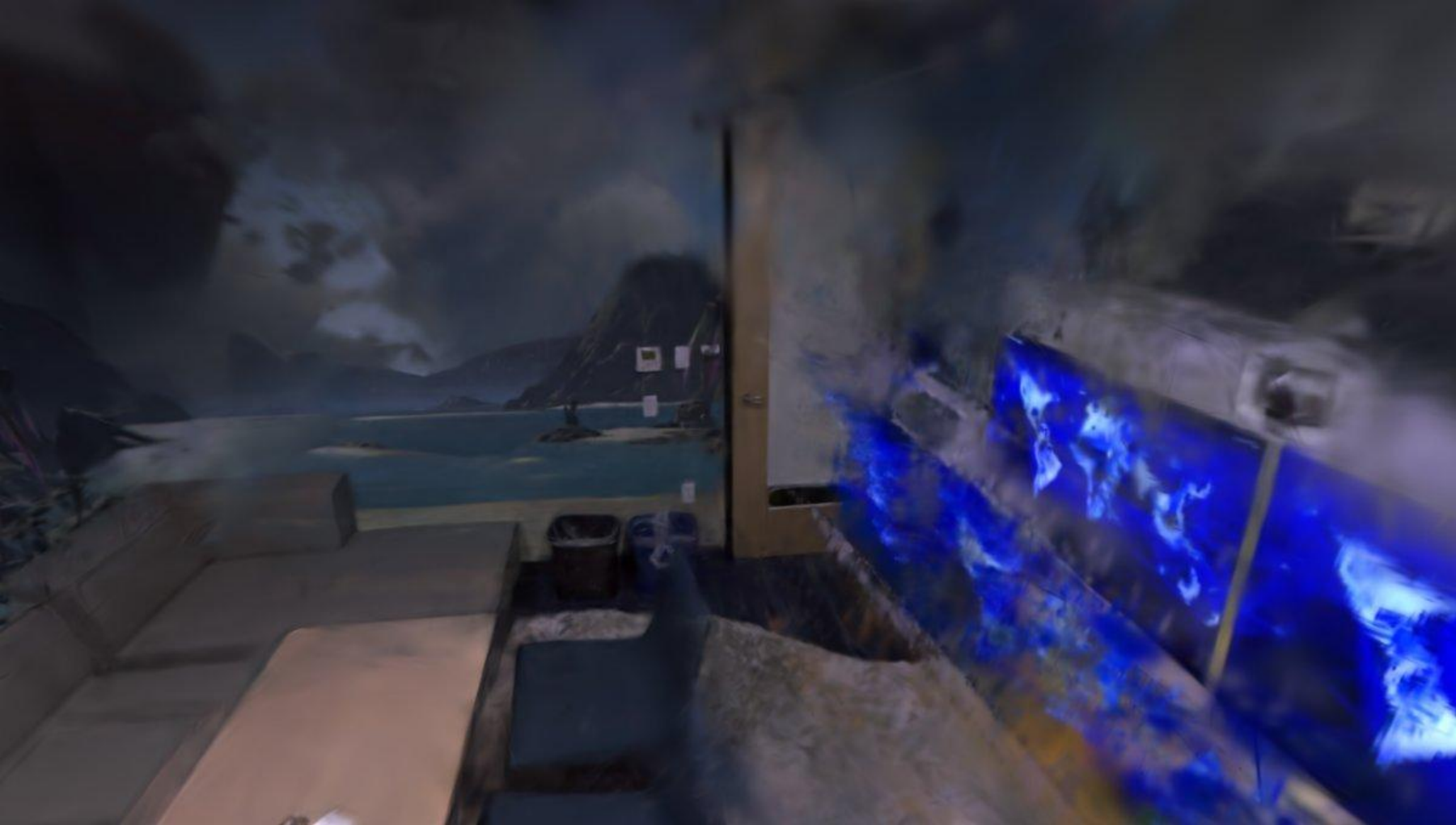} &
        \includegraphics[width=\sz\linewidth]{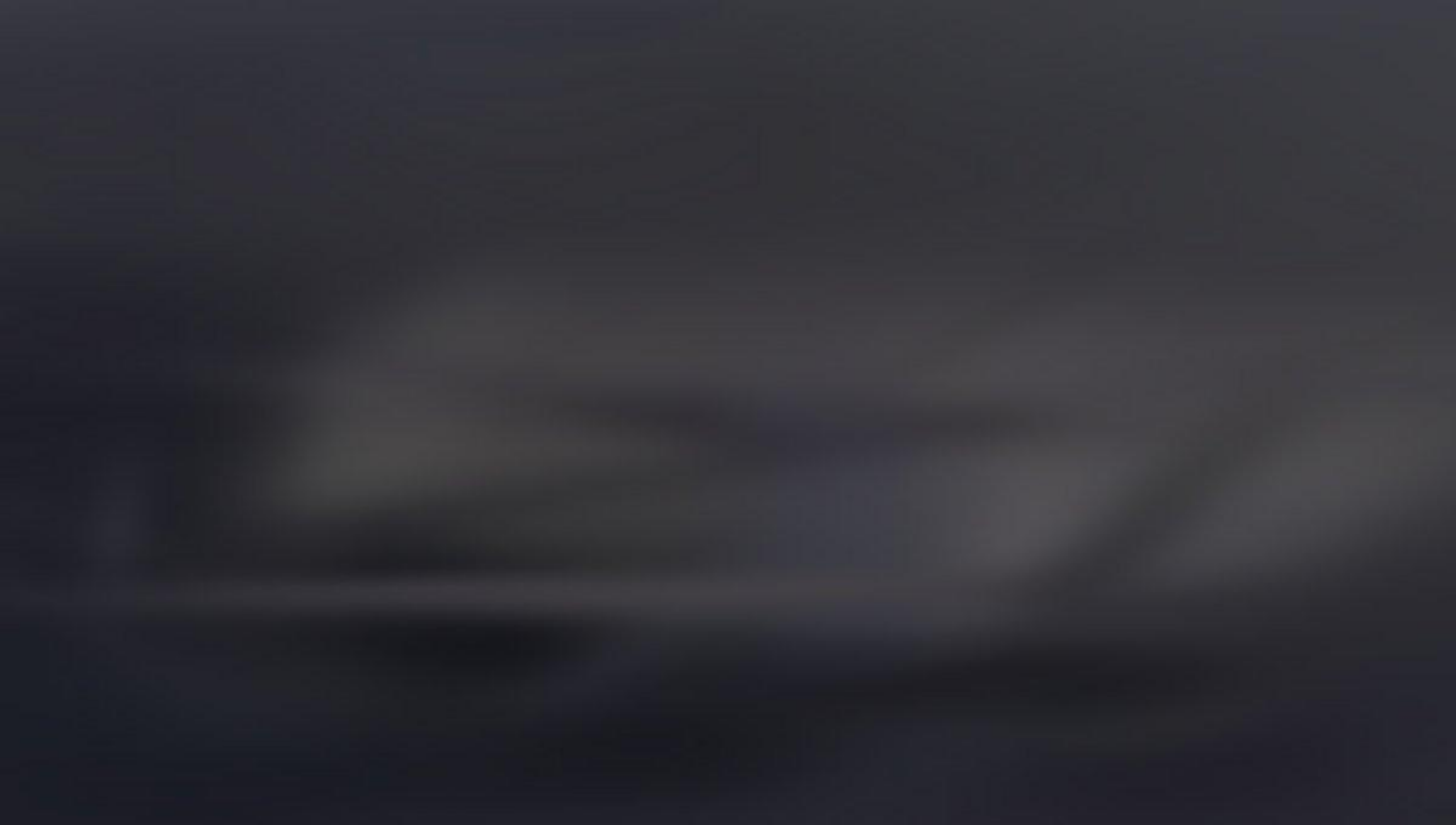} &
        \includegraphics[width=\sz\linewidth]{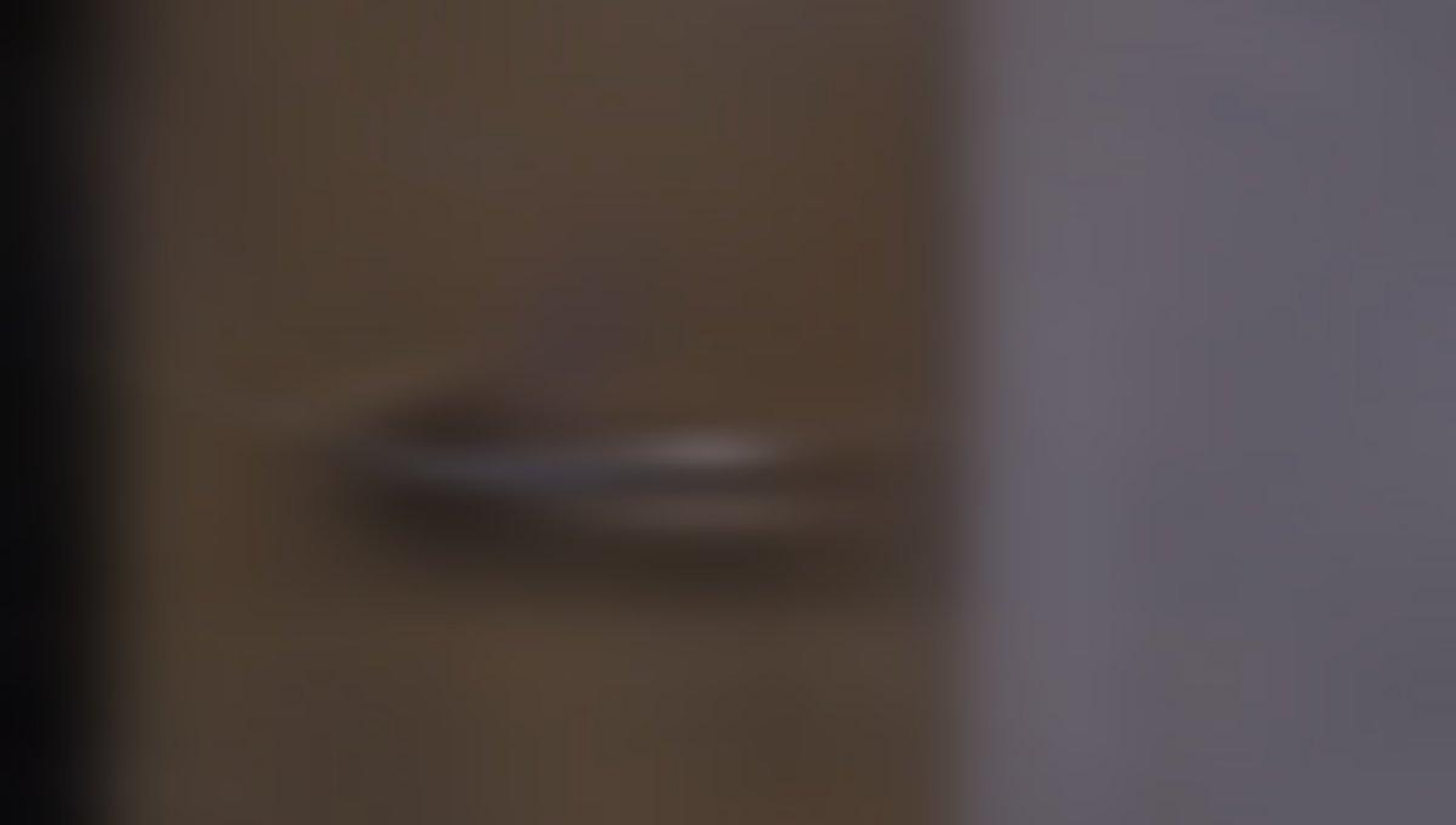}\\

        \rotatebox{90}{\hspace{2pt} GigaSLAM} &
        \includegraphics[width=\sz\linewidth]{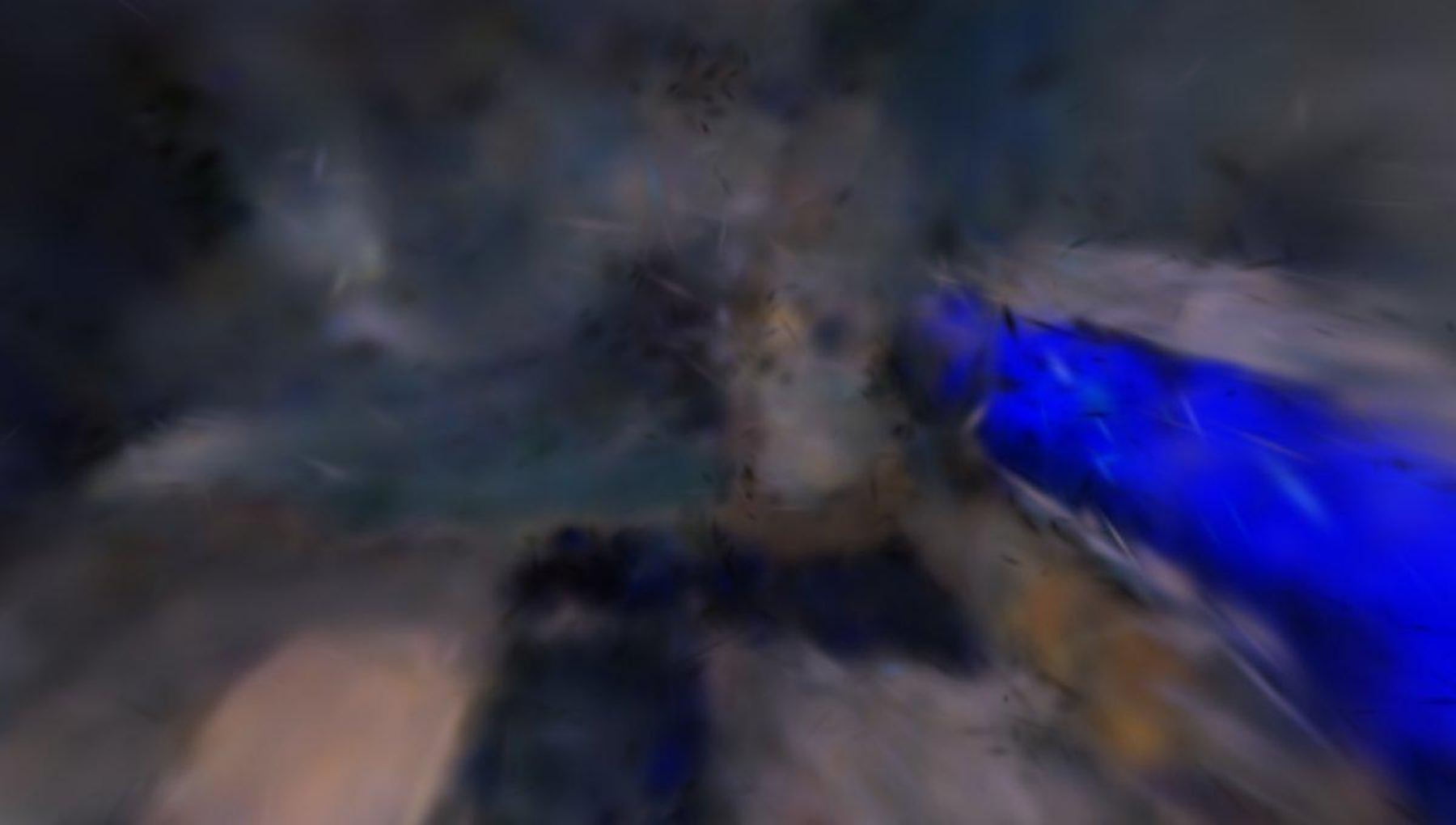} &
        \includegraphics[width=\sz\linewidth]{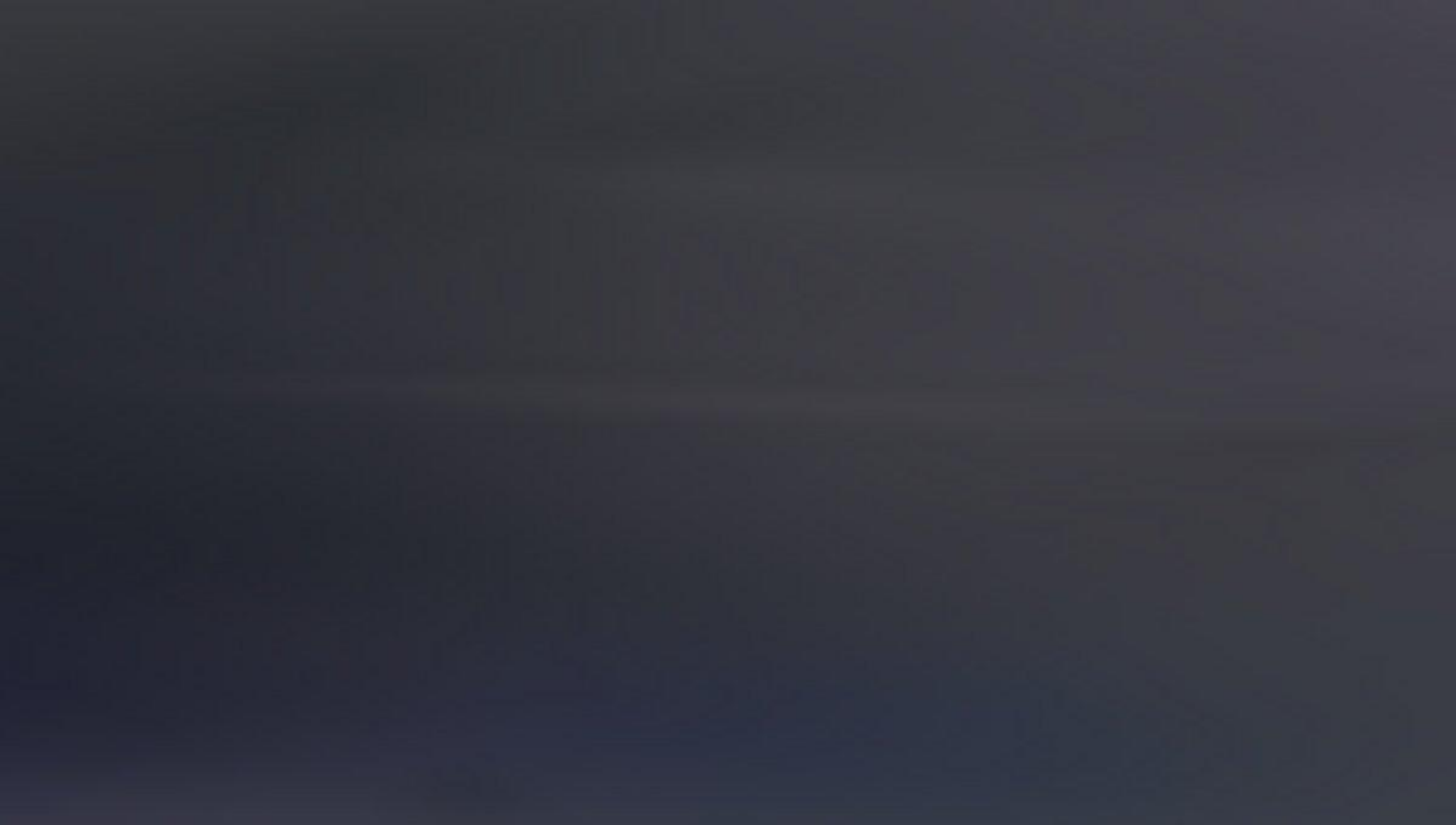} &
        \includegraphics[width=\sz\linewidth]{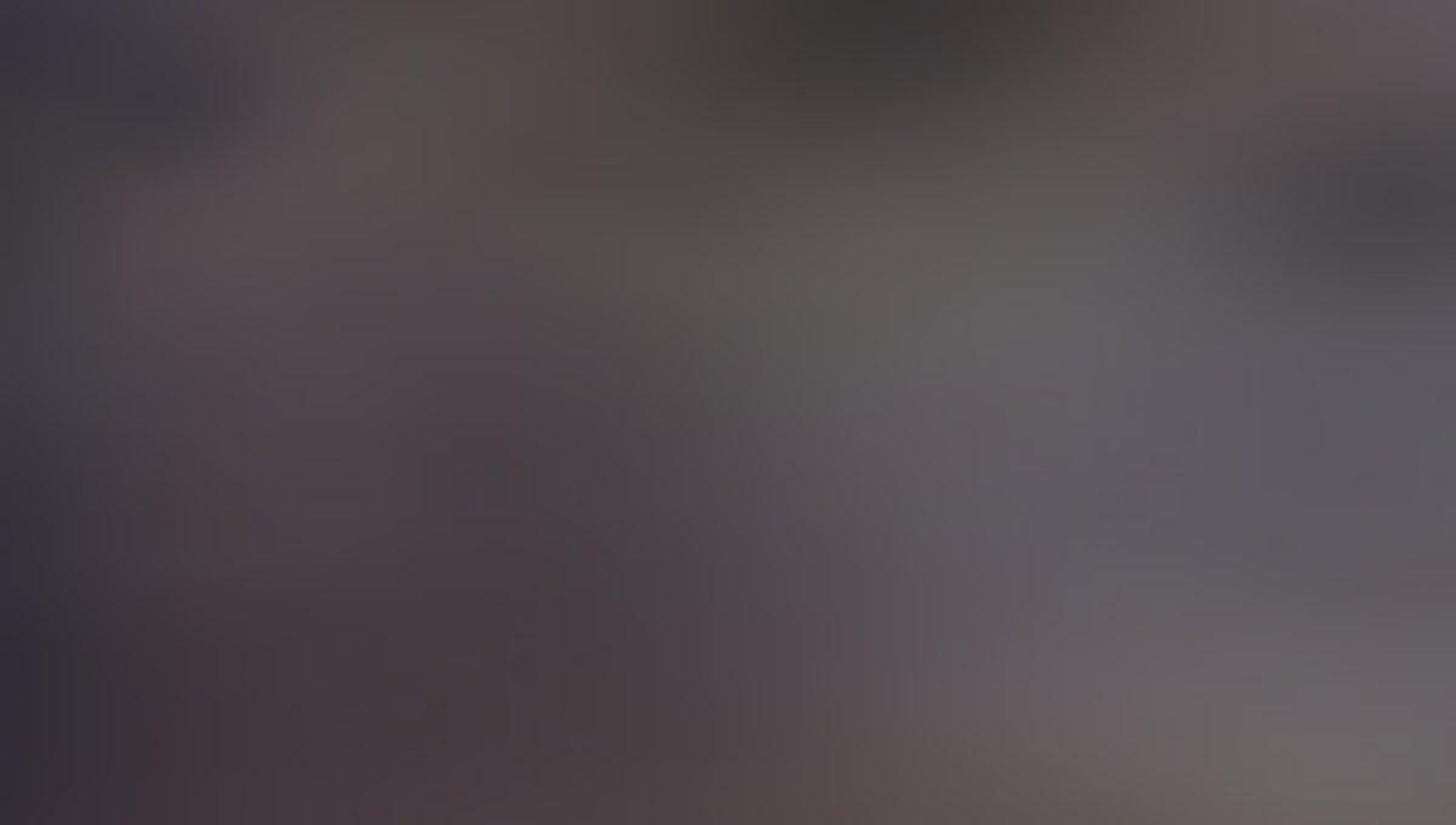}\\

        \rotatebox{90}{\hspace{12pt} \textbf{Ours}} &
        \includegraphics[width=\sz\linewidth]{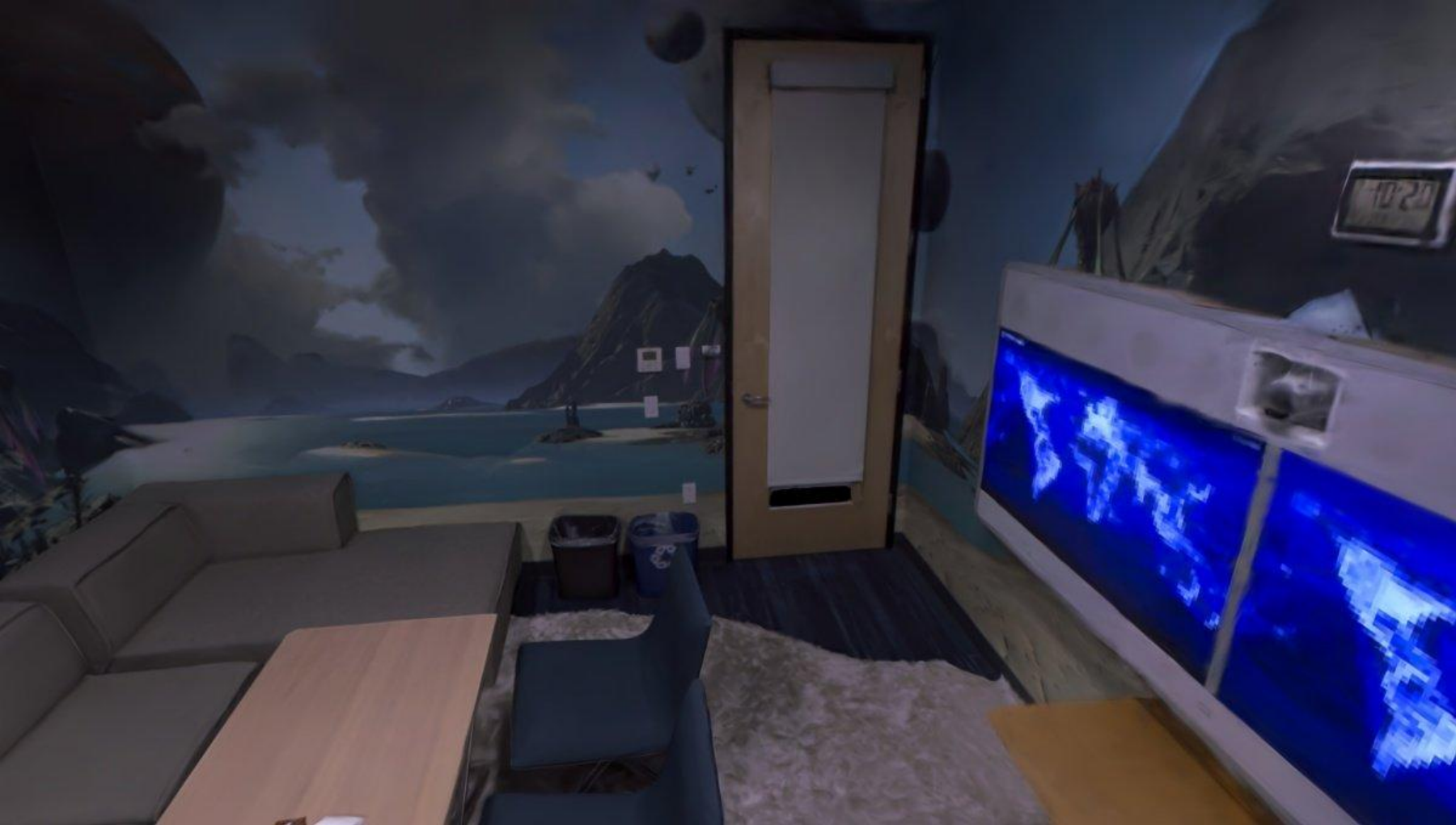} &
        \includegraphics[width=\sz\linewidth]{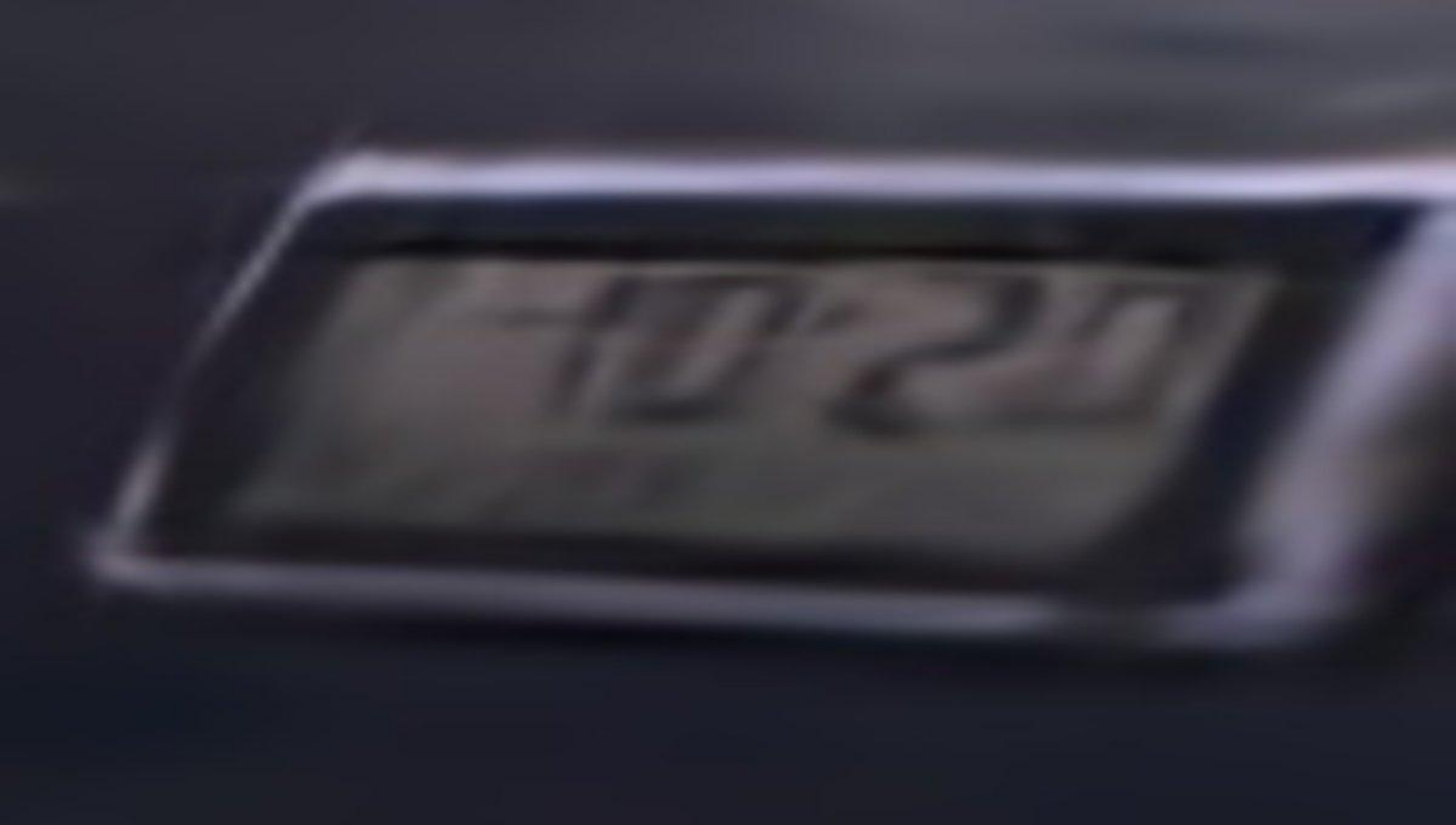} &
        \includegraphics[width=\sz\linewidth]{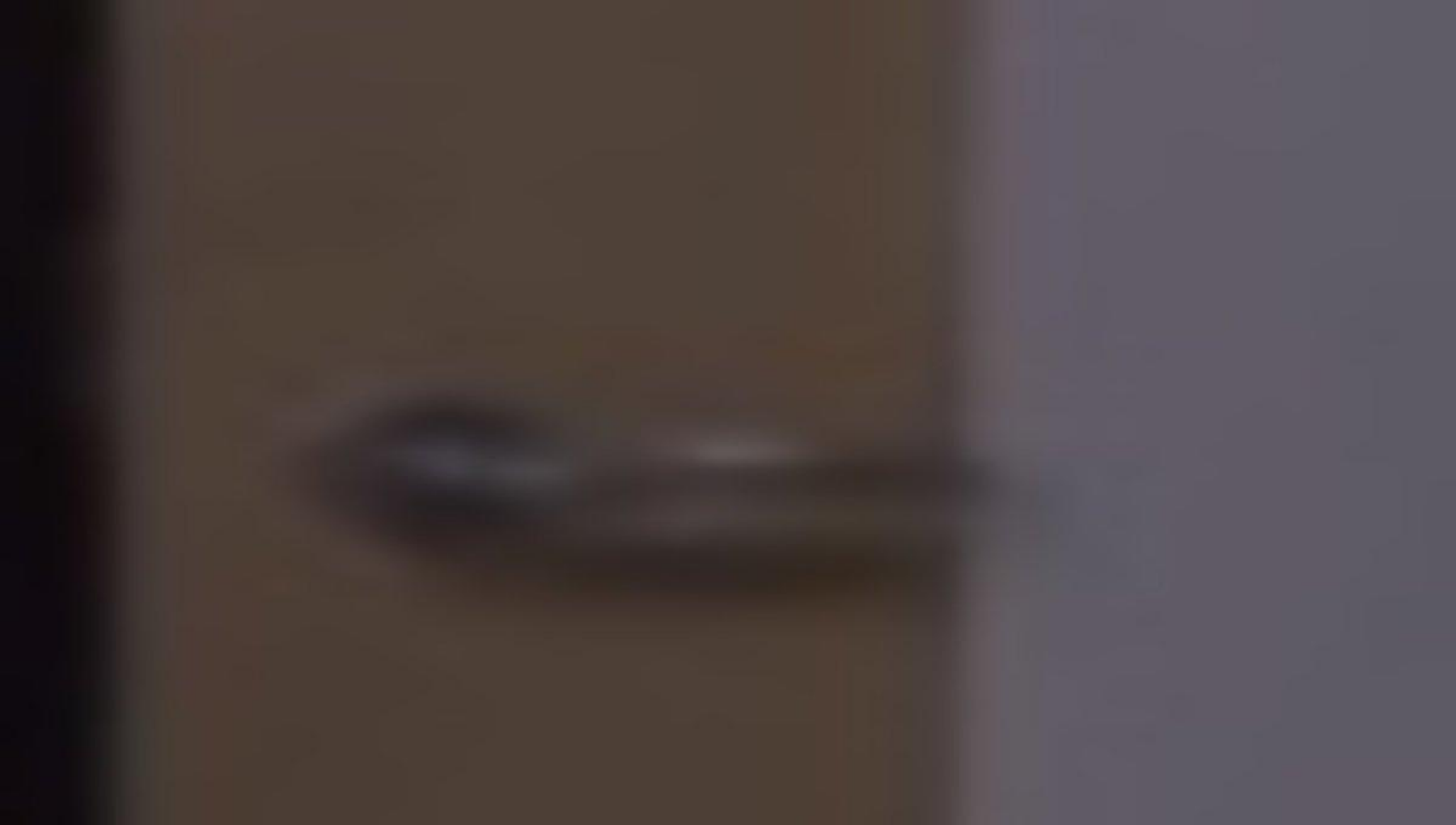}\\
    \end{tabular} 
    \caption{\textbf{Zoomed qualitative results on the Replica~\cite{straub2019replicadatasetdigitalreplica} dataset (office0).}}
    \label{fig:replica_qualitative}
\end{figure}

\subsection{{Datasets and Metrics}}
While most current \ac{3dgs} \ac{slam} methods are limited to small-scale indoor environments due to memory constraints, our approach enables evaluation on long-sequence outdoor scenarios. To demonstrate the versatility of our method across different settings, we evaluate on both established indoor datasets and outdoor sequences. More specifically, we evaluate our method on datasets Replica~\cite{straub2019replicadatasetdigitalreplica}, TUM-RGBD~\cite{sturm12iros}, and KITTI~\cite{Geiger2012CVPR}. Replica~\cite{straub2019replicadatasetdigitalreplica} is a high-quality indoor dataset featuring photorealistic 3D reconstructions of various room environments. TUM-RGBD~\cite{sturm12iros} is a dataset containing RGB-D sequences captured in different indoor environments (offices, hallways, households) with handheld cameras. KITTI~\cite{Geiger2012CVPR} is a comprehensive multi-kilometer outdoor dataset collected from a moving vehicle that includes stereo images, LiDAR scans, and GPS measurements. 

We evaluate localization accuracy using \ac{ate}, and assess reconstruction quality through \ac{psnr}, \ac{ssim}, and \ac{lpips}. For computational efficiency, we measure the total wall-clock time required to process each complete sequence, which provides the most transparent measure of real-world deployment performance. We report this as processing FPS (total frames ÷ total time) to enable fair comparisons across methods that process identical frame sequences, independent of their internal optimization strategies. Note that datasets with lower capture rates (e.g., KITTI at 10 FPS) naturally yield lower processing FPS values due to fewer total frames, while larger-scale outdoor sequences incur additional I/O overhead from chunk management compared to indoor datasets.

\subsection{{Implementation Details}}
We conduct all experiments on an AMD Ryzen 5950X CPU, 64 GB of RAM, PCIe Gen 4 NVMe SSD, and an Nvidia RTX 3090 GPU (24 GB VRAM). Since CaRtGS~\cite{Feng_2025}, GigaSLAM~\cite{deng2025gigaslamlargescalemonocularslam}, and On-The-Fly~\cite{Meuleman_2025} originally reported results on different hardware, we rerun these methods using their public code on our hardware for fair comparison. We acknowledge that some methods require GPUs with greater \ac{vram} capacity than our current hardware; however, our disk-based approach is not constrained by this memory limitation. To ensure fair comparison of online processing capabilities, we evaluate all methods using their online-only performance. For GigaSLAM~\cite{deng2025gigaslamlargescalemonocularslam}, this means excluding its offline post-processing stage. We note that GigaSLAM's~\cite{deng2025gigaslamlargescalemonocularslam} published results include extensive offline refinement, which can achieve higher quality but requires substantial additional computation time, commonly an hour or more per sequence.

CaRtGS~\cite{Feng_2025} and On-The-Fly~\cite{Meuleman_2025} have not been previously evaluated on KITTI~\cite{Geiger2012CVPR} by their original authors. For On-The-Fly~\cite{Meuleman_2025}, we use the default hyperparameters provided by the authors, which are designed to generalize across datasets. For CaRtGS~\cite{Feng_2025}, initially intended for indoor scenes, we fine-tune hyperparameters to optimize performance on outdoor sequences. Our method uses a chunk size of 10 meters for all datasets.

Our method adapts to available processing time, using any time between incoming frames for map refinement. To demonstrate this adaptability, we evaluate at vehicle speeds of 5 km/h and 20 km/h on the KITTI~\cite{Geiger2012CVPR} dataset, corresponding to longer and shorter inter-frame intervals. CaRtGS~\cite{Feng_2025} operates similarly and is evaluated at 5 km/h, while other baseline methods use their default configurations with fixed iteration counts per frame.

\subsection{{Results}}
Quantitative results for KITTI~\cite{Geiger2012CVPR} are shown in Tables \ref{tab:kitti_tracking_ate}, \ref{tab:kitti_tracking_fps} and \ref{tab:kitti_rendering}.
Our method is the only approach that successfully reconstructs all 11 sequences without tracking loss or memory crashes at both intake rates. In contrast, CaRtGS~\cite{Feng_2025} and GigaSLAM~\cite{deng2025gigaslamlargescalemonocularslam} crash on 3 and 4 sequences, respectively, due to VRAM limitations. In comparison, On-the-fly~\cite{Meuleman_2025} loses tracking on 6 sequences due to insufficient tracking robustness and a lack of loop closure capabilities. Where direct comparisons are possible, our method consistently achieves superior reconstruction quality, ranking first or second across most metrics, with particular strength in perceptual quality (LPIPS).
At the 20 km/h intake rate, our method achieves higher frame rates with graceful quality degradation compared to 5 km/h results, demonstrating effective scalability without algorithmic modifications. While On-the-fly~\cite{Meuleman_2025} is faster when it succeeds, its frequent tracking failures limit practical applicability. GigaSLAM~\cite{deng2025gigaslamlargescalemonocularslam}, evaluated in online-only mode as discussed in our implementation details, shows reduced visual quality without its offline refinement stage. These quantitative advantages translate to realistic large-scale reconstructions as visualized in Fig. \ref{fig:qualitative_kitti}. We note that our comparison involves different input modalities (mono vs. stereo), reflecting each method's design choices for handling challenging outdoor scenarios. However, we believe this difference is secondary to the fundamental scalability challenges we address, which occur regardless of input modality.

We observe that our method achieves excellent tracking and visual quality results on the Replica~\cite{straub2019replicadatasetdigitalreplica} dataset, demonstrating strong generalization across both small- and large-scale scenes. Our approach achieves the highest average processing FPS (32.41) on Replica~\cite{straub2019replicadatasetdigitalreplica} among compared methods, as shown in Table \ref{tab:replica_tum_results}. More importantly, we achieve the best reconstruction quality on Replica~\cite{straub2019replicadatasetdigitalreplica} across all visual metrics (PSNR: 34.59, SSIM: 0.94, LPIPS: 0.05), with the superior visual quality evident in Fig.~\ref{fig:replica_qualitative}, where zoomed-in renders reveal reduced artifacts and finer detail preservation compared to baseline methods. On the TUM~\cite{sturm12iros} dataset, our method achieves competitive results, ranking second across all metrics while maintaining efficient runtime.

The Pareto curves in Figure \ref{fig:pareto_comparison} reveal the complete quality-time trade-off spectrum that single-point comparisons in traditional result tables cannot capture. We evaluate on KITTI~\cite{Geiger2012CVPR} scenes 06, 07, and 10, as these are the only sequences where all competing methods complete processing without crashes or tracking failures. We evaluate performance above 1 FPS to focus on methods suitable for online operation. Our method (orange) consistently dominates this trade-off across all three scenes, achieving superior LPIPS scores in significantly less time than competing approaches.

Table \ref{tab:ablation_results} validates our design choices on the KITTI~\cite{Geiger2012CVPR} dataset. Chunking proves essential; omitting this component causes 6 of 11 sequences to fail due to memory constraints. Removing spatial grid keyframe selection causes 2 failures from excessive chunk thrashing. Depth supervision improves perceptual quality (LPIPS: 0.27 vs. 0.28) and structural similarity (SSIM: 0.73 vs. 0.72) while maintaining full sequence completion. These results confirm that chunking enables scalability, grid-based selection ensures efficient I/O, and depth supervision enhances reconstruction quality.

On KITTI~\cite{Geiger2012CVPR} scene 02, the longest sequence at 5.07 km, loading a chunk from disk to VRAM takes an average of 2.8 ms on our NVMe SSD, while saving a chunk requires 2.0 ms. Over the sequence, chunk I/O operations consume 5.5\% of total processing time at the 5 km/h intake rate, completing 44,103 load and 47,999 chunk save operations. At 20 km/h, I/O overhead increases to 19.6\% since I/O costs remain similar while total processing time decreases with less optimization per frame. Disk storage remains practical across all sequences, ranging from 0.3 GB (scene 04) to 7.2 GB (scene 02), with an average of 2.3 GB per sequence, demonstrating that algorithmic chunking with modern NVMe SSDs is a viable alternative to hardware-constrained scaling.

\begin{table}[tb]
\centering
\caption{\textbf{Comparison on TUM~\cite{sturm12iros} and Replica~\cite{straub2019replicadatasetdigitalreplica} (\ac{ate} in cm).} \fscap{Best} and \ndcap{second best} results highlighted.}
\footnotesize
\setlength{\tabcolsep}{8.5pt}
    \begin{tabular}{lcccc}
        \toprule
        \textbf{Metric} & CaRtGS & On-The-Fly & GigaSLAM & \textbf{Ours} \\
        \midrule
        \multicolumn{5}{c}{\textbf{TUM}} \\
        \midrule
        ATE $\downarrow$ & \fs{0.85} & 7.12 & 22.02 & \nd{1.88} \\
        PSNR $\uparrow$ & 20.59 & \fs{23.28} & 13.40 & \nd{21.21} \\
        SSIM $\uparrow$ & 0.71 & \fs{0.84} & 0.53 & \nd{0.75} \\
        LPIPS $\downarrow$ & \nd{0.22} & \fs{0.14} & 0.68 & \nd{0.22} \\
        FPS $\uparrow$ & 24.14 & \fs{51.88} & 10.99 & \nd{25.09} \\
        \midrule
        \multicolumn{5}{c}{\textbf{Replica}} \\
        \midrule
        ATE $\downarrow$ & \fs{0.30} & 31.68 & 11.54 & \nd{0.33} \\
        PSNR $\uparrow$ & \nd{33.85} & 26.91 & 21.11 & \fs{34.59} \\
        SSIM $\uparrow$ & \fs{0.94} & \nd{0.89} & 0.78 & \fs{0.94} \\
        LPIPS $\downarrow$ & \nd{0.07} & 0.16 & 0.38 & \fs{0.05} \\
        FPS $\uparrow$ & \nd{27.65} & 22.96 & 5.65 & \fs{32.41} \\
        \bottomrule
    \end{tabular}
  \label{tab:replica_tum_results}
\end{table}

\begin{table}[tb]
\caption{\textbf{Component Ablation: Success Rate and Quality}}
\footnotesize
\centering
\setlength{\tabcolsep}{5.0pt}
    \begin{tabular}{l|cc|ccc}
        \toprule
        \textbf{Configuration} & \textbf{Success} & \textbf{Failed} & \textbf{PSNR↑} & \textbf{SSIM↑} & \textbf{LPIPS↓} \\
        \midrule
        w/o Chunks & 5 & \textcolor{red}{6} &  \textbf{--} &  \textbf{--} &  \textbf{--} \\
        w/o Grid KF & 9 & \textcolor{red}{2} & \textbf{--} &  \textbf{--} &  \textbf{--} \\
        w/o Depth Sup. & 11 & 0 & 20.90 &  0.72 & 0.28 \\
        \rowcolor{gray!15}
        Full Method & 11 & 0 & 20.89 & 0.73 & 0.27 \\
        \bottomrule
    \end{tabular}
  \label{tab:ablation_results}
\end{table}

To demonstrate our chunking system's effect on VRAM consumption, we plot the allocated VRAM usage for KITTI~\cite{Geiger2012CVPR} scene 02 in Figure \ref{fig:vram_usage}. The initial rise reflects the growing number of Gaussians and keyframes being added to the active set. Once the system reaches the Gaussian budget limit of 1.5 million (lower dashed line), our LRU-based chunk eviction mechanism begins saving older chunks to disk. Similarly, when the number of keyframes reaches 400 (upper dashed line), our disk-based management system maintains only spatially relevant keyframes in memory. As evidenced by the plateau in VRAM usage beyond these thresholds, our system maintains constant memory consumption as the total number of Gaussians and keyframes continues to grow throughout the sequence.

We evaluate our method on Nvidia's Jetson AGX Orin, an edge AI computing platform with energy-efficient capabilities suitable for robotics, drones, and other mobile computing scenarios. Quantitative results in Table \ref{tab:jetson_results} show that our method maintains competitive performance on the resource-constrained platform with only modest degradation in rendering quality and increased processing time, demonstrating practical viability for mobile applications.

\begin{figure}[htbp]
\centering
\begin{tikzpicture}[scale=0.65]
\pgfplotsset{
    scale only axis,
    % scaled x ticks=base 10:3,
    xmin=0, xmax=3500
}

% First axis (left y-axis)
\begin{axis}[
  axis y line*=left,
  ymin=0, ymax=7,
  height=7cm,
  xlabel=\text{Time [s]},
  ylabel=\text{VRAM Allocated [GB]},
  legend pos=south east,
  legend style={font=\footnotesize}
]
\addplot[smooth,blue,line width=2pt]
  coordinates{
    (17.336,0.338138)(23.655,0.43042)(30.651,0.536223)(37.187,0.643977)(44.164,0.751905)(51.227,0.879752)(58.681,1.00148)(65.294,1.11907)(71.871,1.2101)(78.969,1.34663)(86.445,1.47987)(94.046,1.609)(101.331,1.73964)(108.928,1.86903)(117.28,2.02957)(125.847,2.17529)(132.737,2.28971)(139.621,2.4117)(145.8,2.53744)(151.969,2.62263)(158.651,2.68715)(166.044,2.74952)(172.849,2.8618)(181.123,3.00355)(189.151,3.12828)(197.328,3.27733)(204.902,3.41226)(212.604,3.54174)(220.714,3.68838)(228.913,3.84205)(237.699,3.98276)(246.038,4.12662)(254.367,4.27454)(262.747,4.42723)(271.259,4.57875)(280.232,4.73926)(289.169,4.8966)(298.363,5.06486)(307.903,5.22291)(317.023,5.39091)(326.317,5.54396)(334.75,5.68412)(343.127,5.82768)(350.932,5.87142)(358.515,5.8889)(365.678,5.90375)(373.398,5.92029)(381.581,5.93897)(389.885,5.95382)(397.898,5.96438)(407.137,5.98112)(415.325,5.99148)(423.533,6.01082)(432.649,6.04782)(441.135,6.0493)(449.639,6.0634)(458.819,6.06061)(467.038,6.07174)(475.391,6.08469)(484.018,6.12811)(492.85,6.1282)(502.224,6.1465)(511.679,6.16964)(520.821,6.18604)(529.959,6.20949)(539.199,6.22648)(548.517,6.23769)(558.748,6.27503)(569.198,6.24397)(580.001,6.2657)(591.912,6.23058)(603.067,6.27138)(614.468,6.25379)(625.939,6.23299)(635.994,6.25916)(644.307,6.27965)(652.905,6.24781)(662.365,6.27353)(672.551,6.25639)(681.858,6.26708)(691.337,6.23244)(700.784,6.24576)(711.571,6.26925)(721.658,6.23405)(732.288,6.25745)(743.134,6.27974)(753.667,6.23495)(762.295,6.24219)(770.198,6.24576)(778.309,6.26185)(788.879,6.27387)(799.672,6.23556)(810.168,6.25269)(819.412,6.26488)(827.341,6.27906)(835.596,6.28842)(844.125,6.24483)(852.528,6.24883)(860.472,6.25168)(868.516,6.25819)(876.565,6.26862)(884.565,6.27332)(893.034,6.23419)(901.481,6.25175)(910.18,6.27015)(919.39,6.23945)(928.592,6.27717)(938.694,6.28541)(950.471,6.25583)(961.806,6.28271)(973.024,6.26916)(983.983,6.28062)(995.079,6.2462)(1006.18,6.25902)(1017.38,6.27849)(1028.82,6.24112)(1040.15,6.26555)(1051.73,6.27069)(1063.63,6.28505)(1075.35,6.2403)(1086.99,6.266)(1098.56,6.2768)(1110,6.27001)(1121.79,6.27825)(1133.59,6.2471)(1145.21,6.27802)(1153.44,6.28195)(1161.69,6.29758)(1170.13,6.26189)(1178.99,6.27873)(1188.3,6.29807)(1197.79,6.25933)(1207.27,6.26596)(1217.25,6.28305)(1227.2,6.29689)(1237.12,6.25493)(1246.98,6.26933)(1256.9,6.28567)(1266.9,6.30652)(1277.54,6.27511)(1288.01,6.28647)(1298.63,6.30531)(1307.38,6.26976)(1315.68,6.28371)(1324.35,6.29442)(1333.21,6.26404)(1342.3,6.27536)(1351.74,6.28711)(1361.35,6.31735)(1371.04,6.25842)(1380.92,6.27165)(1391.34,6.28376)(1402.48,6.31561)(1413.22,6.26501)(1423.9,6.28381)(1437.23,6.30333)(1448.86,6.26818)(1460.72,6.28538)(1472.33,6.30209)(1483.23,6.26283)(1493.3,6.26303)(1503.26,6.26718)(1513.61,6.27287)(1523.23,6.27755)(1531.34,6.28312)(1539.95,6.29452)(1548.82,6.26442)(1557.98,6.27977)(1567.94,6.29179)(1578.51,6.2646)(1589.26,6.275)(1597.83,6.29722)(1606.13,6.27989)(1616.71,6.26823)(1627.08,6.25362)(1639.22,6.24645)(1652.66,6.24026)(1666.25,6.28091)(1680.28,6.29679)(1692.84,6.26542)(1705.43,6.28245)(1717.53,6.29469)(1730.51,6.27681)(1741.81,6.28664)(1754.3,6.26484)(1765.85,6.28158)(1778.24,6.31894)(1790.62,6.27547)(1804.14,6.3109)(1818.98,6.28878)(1832.16,6.26593)(1844.38,6.2854)(1856.83,6.26154)(1870.22,6.29371)(1882.7,6.27228)(1895.07,6.25058)(1908.22,6.25597)(1921.71,6.26838)(1937.24,6.27256)(1949.2,6.24846)(1960.56,6.27273)(1971.97,6.28147)(1983.62,6.25855)(1994.85,6.27586)(2006.42,6.2529)(2018.01,6.26825)(2029.68,6.2741)(2041.49,6.28376)(2053.07,6.24433)(2064.34,6.25028)(2075.7,6.25492)(2087.72,6.27778)(2099.86,6.27711)(2111.41,6.26385)(2123.16,6.27974)(2134.05,6.29379)(2142.21,6.25205)(2151.12,6.25739)(2160.3,6.26348)(2168.7,6.27401)(2176.94,6.27848)(2185.12,6.23899)(2193.33,6.25111)(2201.79,6.26132)(2211.19,6.27694)(2222.18,6.2879)(2231.99,6.25995)(2241.04,6.26702)(2250.51,6.28653)(2260.03,6.2459)(2269.58,6.25932)(2279.25,6.27653)(2289.46,6.28992)(2300.72,6.25169)(2311.6,6.26756)(2322.18,6.28097)(2333.25,6.26767)(2342.14,6.25805)(2350.1,6.26696)(2358.95,6.27504)(2368.02,6.28765)(2377.35,6.25793)(2386.71,6.264)(2398.03,6.23429)(2408.02,6.25299)(2417.86,6.26332)(2428.43,6.27869)(2438.92,6.24599)(2449.83,6.27232)(2462.03,6.24101)(2473.37,6.26104)(2484.43,6.27322)(2495.06,6.24815)(2503.93,6.25361)(2513.06,6.26711)(2523.04,6.23963)(2531.83,6.251)(2541.99,6.26349)(2551.51,6.2774)(2560.82,6.28099)(2570.12,6.243)(2579.27,6.25849)(2589.12,6.27605)(2598.99,6.24037)(2609.07,6.26474)(2620.45,6.28406)(2631.87,6.24517)(2641.58,6.2625)(2650.12,6.27418)(2659.24,6.24513)(2667.45,6.26033)(2675.92,6.26984)(2685.01,6.23704)(2695.4,6.25369)(2705.68,6.27539)(2716.17,6.25294)(2726.92,6.27335)(2737.69,6.23545)(2748.55,6.25598)(2766.18,6.28117)(2779.65,6.27496)(2792.72,6.27067)(2807.39,6.23227)(2822.13,6.25082)(2838.57,6.28729)(2857.49,6.2343)(2873.83,6.24146)(2888.92,6.25777)(2902.57,6.25629)(2910.56,6.26688)(2918.86,6.23471)(2927.14,6.24756)(2935.56,6.2552)(2944.18,6.26501)(2953.91,6.23009)(2963.03,6.24656)(2972.44,6.27038)(2982.75,6.23938)(2993.26,6.24959)(3002.46,6.26726)(3010.62,6.23128)(3018.72,6.2442)(3027.34,6.26322)(3036.62,6.26697)(3046.35,6.23434)(3056.09,6.24627)(3066.4,6.25791)(3077.27,6.22511)(3092.81,6.27059)(3107.26,6.24314)(3120.5,6.26029)(3133.56,6.26921)(3146.39,6.22163)(3159.8,6.26195)(3171.08,6.25592)(3179.94,6.265)(3188.72,6.27664)(3197.48,6.23497)(3205.95,6.24509)(3214.4,6.25698)(3223.07,6.26802)(3231.98,6.28203)(3241.23,6.23993)(3250.31,6.24784)(3259.63,6.26674)(3269.29,6.27719)(3278.94,6.23151)(3288.63,6.24845)(3298.74,6.2771)(3308.64,6.24466)(3316.93,6.264)(3325.52,6.27891)(3334.18,6.23277)(3342.65,6.24781)(3351.66,6.26487)(3360.89,6.27641)(3370.59,6.24057)(3380.16,6.25468)(3390.07,6.27534)(3400.24,6.24063)(3410.47,6.25002)(3421.01,6.25262)(3431.43,6.25669)(3440.99,6.26787)(3450.99,6.25765)(3461.26,6.25933)(3472.11,6.24367)(3483.63,6.24436)(3496.99,6.25454)(3510.6,6.25)(3525.85,6.24634)(3606.26,5.14463)(3655.2,6.24694)(3666.87,6.22284)(3677.29,6.23198)(3688.27,6.24168)(3699.65,6.23658)(3710.97,6.25978)(3723.71,6.25333)(3737.49,6.22676)(3751.69,6.22738)(3766.47,6.24862)(3781.51,6.26101)(3835.11,2.96813)(3850.25,4.59409)(3866.97,6.12392)(3881.22,6.1311)(3896.69,6.14026)(3930.48,6.28772)(3970.38,6.1907)(4003.24,6.20375)(4041.88,6.18805)(4055.59,6.2084)
}; \label{Hplot}
\end{axis}

% Second axis (right y-axis)
\begin{axis}[
  axis y line*=right,
  axis x line=none,
  ymin=0, ymax=9,
  height=7cm,
  ylabel=Keyframes / Gaussians,
  legend pos=south east,
  ytick={1, 2, 3, 4, 5, 6, 7, 8, 9},
  yticklabels={100/1M, 200/2M, 300/3M, 400/4M, 500/5M, 600/6M, 700/7M, 800/8M, 900/9M},
]
% Add the first plot to the legend using its reference
\addlegendimage{/pgfplots/refstyle=Hplot}\addlegendentry{\text{VRAM [GB]}}

% Add the second plot
\addplot[smooth,red,line width=2pt]
  coordinates{
    (17.336,0.061284)(23.655,0.066516)(30.651,0.074439)(37.187,0.084241)(44.164,0.09464)(51.227,0.115003)(58.681,0.132014)(65.294,0.156384)(71.871,0.156893)(78.969,0.182122)(86.445,0.203438)(94.046,0.21939)(101.331,0.2436)(108.928,0.261378)(117.28,0.3047)(125.847,0.328544)(132.737,0.347762)(139.621,0.37035)(145.8,0.397569)(151.969,0.413697)(158.651,0.433485)(166.044,0.454016)(172.849,0.471638)(181.123,0.489263)(189.151,0.50153)(197.328,0.525348)(204.902,0.541095)(212.604,0.553709)(220.714,0.577659)(228.913,0.612851)(237.699,0.644938)(246.038,0.664649)(254.367,0.687025)(262.747,0.709278)(271.259,0.733233)(280.232,0.755924)(289.169,0.780409)(298.363,0.810744)(307.903,0.827506)(317.023,0.862557)(326.317,0.881399)(334.75,0.899998)(343.127,0.924507)(350.932,0.949283)(358.515,0.973352)(365.678,0.991353)(373.398,1.009827)(381.581,1.03587)(389.885,1.05701)(397.898,1.065627)(407.137,1.086368)(415.325,1.103055)(423.533,1.122916)(432.649,1.151054)(441.135,1.1665)(449.639,1.182804)(458.819,1.184646)(467.038,1.199098)(475.391,1.221421)(484.018,1.264186)(492.85,1.275793)(502.224,1.302647)(511.679,1.333506)(520.821,1.358467)(529.959,1.389834)(539.199,1.4104)(548.517,1.432113)(558.748,1.48399)(569.198,1.514585)(580.001,1.544974)(591.912,1.577488)(603.067,1.632073)(614.468,1.690597)(625.939,1.734817)(635.994,1.775859)(644.307,1.803654)(652.905,1.844984)(662.365,1.884397)(672.551,1.937925)(681.858,1.955323)(691.337,1.980189)(700.784,2.001451)(711.571,2.036402)(721.658,2.068292)(732.288,2.101527)(743.134,2.134529)(753.667,2.148196)(762.295,2.16105)(770.198,2.16625)(778.309,2.188085)(788.879,2.204626)(799.672,2.231161)(810.168,2.258993)(819.412,2.27487)(827.341,2.292494)(835.596,2.308331)(844.125,2.321384)(852.528,2.328227)(860.472,2.331047)(868.516,2.341939)(876.565,2.357183)(884.565,2.365217)(893.034,2.38518)(901.481,2.412058)(910.18,2.438507)(919.39,2.468904)(928.592,2.508016)(938.694,2.536259)(950.471,2.576465)(961.806,2.595955)(973.024,2.59907)(983.983,2.611331)(995.079,2.632147)(1006.18,2.657106)(1017.38,2.680696)(1028.82,2.703147)(1040.15,2.722331)(1051.73,2.746375)(1063.63,2.768963)(1075.35,2.79239)(1086.99,2.812538)(1098.56,2.827296)(1110,2.837735)(1121.79,2.853047)(1133.59,2.879657)(1145.21,2.906882)(1153.44,2.914954)(1161.69,2.937691)(1170.13,2.971628)(1178.99,2.997327)(1188.3,3.027845)(1197.79,3.053634)(1207.27,3.067511)(1217.25,3.088153)(1227.2,3.110064)(1237.12,3.120678)(1246.98,3.144365)(1256.9,3.172104)(1266.9,3.202294)(1277.54,3.239799)(1288.01,3.255426)(1298.63,3.283449)(1307.38,3.305601)(1315.68,3.329558)(1324.35,3.352463)(1333.21,3.379363)(1342.3,3.400849)(1351.74,3.421565)(1361.35,3.44476)(1371.04,3.454982)(1380.92,3.478083)(1391.34,3.496615)(1402.48,3.526816)(1413.22,3.540123)(1423.9,3.572813)(1437.23,3.621521)(1448.86,3.638572)(1460.72,3.671892)(1472.33,3.701756)(1483.23,3.704034)(1493.3,3.709193)(1503.26,3.716918)(1513.61,3.731914)(1523.23,3.739356)(1531.34,3.74852)(1539.95,3.767118)(1548.82,3.793031)(1557.98,3.820186)(1567.94,3.844875)(1578.51,3.869255)(1589.26,3.894045)(1597.83,3.929349)(1606.13,3.979368)(1616.71,4.037263)(1627.08,4.104168)(1639.22,4.181015)(1652.66,4.263317)(1666.25,4.32261)(1680.28,4.348574)(1692.84,4.367487)(1705.43,4.398222)(1717.53,4.419052)(1730.51,4.462534)(1741.81,4.486424)(1754.3,4.523671)(1765.85,4.554744)(1778.24,4.595415)(1790.62,4.623833)(1804.14,4.69586)(1818.98,4.744275)(1832.16,4.804691)(1844.38,4.838452)(1856.83,4.881708)(1870.22,4.938158)(1882.7,4.98494)(1895.07,5.017782)(1908.22,5.084474)(1921.71,5.15077)(1937.24,5.223957)(1949.2,5.261259)(1960.56,5.30006)(1971.97,5.319871)(1983.62,5.348703)(1994.85,5.369384)(2006.42,5.41418)(2018.01,5.439371)(2029.68,5.452037)(2041.49,5.470789)(2053.07,5.486935)(2064.34,5.495997)(2075.7,5.50503)(2087.72,5.529349)(2099.86,5.557175)(2111.41,5.59435)(2123.16,5.615872)(2134.05,5.638513)(2142.21,5.648645)(2151.12,5.656949)(2160.3,5.675934)(2168.7,5.691392)(2176.94,5.700025)(2185.12,5.722087)(2193.33,5.74091)(2201.79,5.756869)(2211.19,5.782869)(2222.18,5.800885)(2231.99,5.829626)(2241.04,5.837899)(2250.51,5.871315)(2260.03,5.886134)(2269.58,5.912018)(2279.25,5.93951)(2289.46,5.96172)(2300.72,5.980345)(2311.6,6.006776)(2322.18,6.02664)(2333.25,6.062855)(2342.14,6.066226)(2350.1,6.083939)(2358.95,6.105881)(2368.02,6.125662)(2377.35,6.157557)(2386.71,6.165501)(2398.03,6.201769)(2408.02,6.235711)(2417.86,6.253793)(2428.43,6.281378)(2438.92,6.308452)(2449.83,6.350011)(2462.03,6.38061)(2473.37,6.413621)(2484.43,6.434704)(2495.06,6.468974)(2503.93,6.476842)(2513.06,6.503337)(2523.04,6.533421)(2531.83,6.55248)(2541.99,6.573053)(2551.51,6.59698)(2560.82,6.606854)(2570.12,6.623915)(2579.27,6.644296)(2589.12,6.673357)(2598.99,6.694716)(2609.07,6.736576)(2620.45,6.767662)(2631.87,6.781695)(2641.58,6.810518)(2650.12,6.831475)(2659.24,6.85748)(2667.45,6.887077)(2675.92,6.903813)(2685.01,6.931545)(2695.4,6.960588)(2705.68,6.98782)(2716.17,7.030652)(2726.92,7.06388)(2737.69,7.090045)(2748.55,7.125479)(2766.18,7.201464)(2779.65,7.216066)(2792.72,7.210138)(2807.39,7.222524)(2822.13,7.254786)(2838.57,7.290775)(2857.49,7.307299)(2873.83,7.318365)(2888.92,7.344963)(2902.57,7.390932)(2910.56,7.409772)(2918.86,7.432447)(2927.14,7.451396)(2935.56,7.463312)(2944.18,7.47998)(2953.91,7.502947)(2963.03,7.525437)(2972.44,7.563585)(2982.75,7.595628)(2993.26,7.618844)(3002.46,7.644777)(3010.62,7.670583)(3018.72,7.690061)(3027.34,7.71868)(3036.62,7.729208)(3046.35,7.750355)(3056.09,7.77008)(3066.4,7.786332)(3077.27,7.826952)(3092.81,7.896917)(3107.26,7.934026)(3120.5,7.96204)(3133.56,7.976555)(3146.39,7.99836)(3159.8,8.038625)(3171.08,8.047973)(3179.94,8.064871)(3188.72,8.082147)(3197.48,8.092621)(3205.95,8.108788)(3214.4,8.12697)(3223.07,8.141332)(3231.98,8.163081)(3241.23,8.183265)(3250.31,8.197227)(3259.63,8.223661)(3269.29,8.239823)(3278.94,8.246483)(3288.63,8.274897)(3298.74,8.317259)(3308.64,8.345606)(3316.93,8.374412)(3325.52,8.39837)(3334.18,8.417536)(3342.65,8.44109)(3351.66,8.468654)(3360.89,8.487729)(3370.59,8.511365)(3380.16,8.534411)(3390.07,8.569383)(3400.24,8.587701)(3410.47,8.599366)(3421.01,8.605469)(3431.43,8.610309)(3440.99,8.626375)(3450.99,8.651951)(3461.26,8.680858)(3472.11,8.713895)(3483.63,8.74627)(3496.99,8.775883)(3510.6,8.802119)(3525.85,8.826818)(3606.26,8.811626)(3655.2,8.827297)(3666.87,8.814532)(3677.29,8.809221)(3688.27,8.84083)(3699.65,8.906868)(3710.97,8.937054)(3723.71,9.021821)(3737.49,9.058958)(3751.69,9.13257)(3766.47,9.226765)(3781.51,9.247713)(3835.11,9.228108)(3850.25,9.225891)(3866.97,9.239347)(3881.22,9.247992)(3896.69,9.260337)(3930.48,9.309904)(3970.38,9.291893)(4003.24,9.295179)(4041.88,9.357205)(4055.59,9.400189)
}; \addlegendentry{Total Gaussians}

\addplot[smooth,green,line width=2pt]
  coordinates{
    (17.336,0.061284)(23.655,0.066516)(30.651,0.074439)(37.187,0.084241)(44.164,0.09464)(51.227,0.115003)(58.681,0.132014)(65.294,0.156384)(71.871,0.156893)(78.969,0.182122)(86.445,0.203438)(94.046,0.21939)(101.331,0.2436)(108.928,0.261378)(117.28,0.3047)(125.847,0.328544)(132.737,0.347762)(139.621,0.37035)(145.8,0.397569)(151.969,0.413697)(158.651,0.433485)(166.044,0.454016)(172.849,0.471638)(181.123,0.489263)(189.151,0.50153)(197.328,0.525348)(204.902,0.541095)(212.604,0.553709)(220.714,0.577659)(228.913,0.612851)(237.699,0.644938)(246.038,0.664649)(254.367,0.687025)(262.747,0.709278)(271.259,0.733233)(280.232,0.755924)(289.169,0.780409)(298.363,0.810744)(307.903,0.827506)(317.023,0.862557)(326.317,0.881399)(334.75,0.899998)(343.127,0.924507)(350.932,0.949283)(358.515,0.973352)(365.678,0.991353)(373.398,1.009827)(381.581,1.03587)(389.885,1.05701)(397.898,1.065627)(407.137,1.086368)(415.325,1.103055)(423.533,1.122916)(432.649,1.151054)(441.135,1.1665)(449.639,1.182804)(458.819,1.184646)(467.038,1.199098)(475.391,1.221421)(484.018,1.264186)(492.85,1.275793)(502.224,1.302647)(511.679,1.333506)(520.821,1.358467)(529.959,1.389834)(539.199,1.4104)(548.517,1.432113)(558.748,1.48399)(569.198,1.436117)(580.001,1.466506)(591.912,1.41556)(603.067,1.470145)(614.468,1.446556)(625.939,1.41316)(635.994,1.454202)(644.307,1.481997)(652.905,1.435748)(662.365,1.475161)(672.551,1.451894)(681.858,1.469292)(691.337,1.413581)(700.784,1.434843)(711.571,1.469794)(721.658,1.417494)(732.288,1.450729)(743.134,1.483731)(753.667,1.417557)(762.295,1.430411)(770.198,1.435611)(778.309,1.457446)(788.879,1.473987)(799.672,1.416217)(810.168,1.444049)(819.412,1.459926)(827.341,1.47755)(835.596,1.493387)(844.125,1.428838)(852.528,1.435681)(860.472,1.438501)(868.516,1.449393)(876.565,1.464637)(884.565,1.472671)(893.034,1.412297)(901.481,1.439175)(910.18,1.465624)(919.39,1.419694)(928.592,1.458806)(938.694,1.487049)(950.471,1.443553)(961.806,1.463043)(973.024,1.466158)(983.983,1.478419)(995.079,1.422532)(1006.18,1.447491)(1017.38,1.471081)(1028.82,1.412969)(1040.15,1.432153)(1051.73,1.456197)(1063.63,1.478785)(1075.35,1.417512)(1086.99,1.43766)(1098.56,1.452418)(1110,1.462857)(1121.79,1.478169)(1133.59,1.428151)(1145.21,1.455376)(1153.44,1.463448)(1161.69,1.486185)(1170.13,1.433547)(1178.99,1.459246)(1188.3,1.489764)(1197.79,1.42629)(1207.27,1.440167)(1217.25,1.460809)(1227.2,1.48272)(1237.12,1.408478)(1246.98,1.432165)(1256.9,1.459904)(1266.9,1.490094)(1277.54,1.442118)(1288.01,1.457745)(1298.63,1.485768)(1307.38,1.431237)(1315.68,1.455194)(1324.35,1.478099)(1333.21,1.426743)(1342.3,1.448229)(1351.74,1.468945)(1361.35,1.49214)(1371.04,1.413911)(1380.92,1.437012)(1391.34,1.455544)(1402.48,1.485745)(1413.22,1.423468)(1423.9,1.456158)(1437.23,1.491339)(1448.86,1.428272)(1460.72,1.461592)(1472.33,1.491456)(1483.23,1.417185)(1493.3,1.422344)(1503.26,1.430069)(1513.61,1.445065)(1523.23,1.452507)(1531.34,1.461671)(1539.95,1.480269)(1548.82,1.424423)(1557.98,1.451578)(1567.94,1.476267)(1578.51,1.422925)(1589.26,1.447715)(1597.83,1.483019)(1606.13,1.450732)(1616.71,1.430909)(1627.08,1.408563)(1639.22,1.400661)(1652.66,1.397973)(1666.25,1.457266)(1680.28,1.48323)(1692.84,1.423693)(1705.43,1.454428)(1717.53,1.475258)(1730.51,1.436019)(1741.81,1.459909)(1754.3,1.418528)(1765.85,1.449601)(1778.24,1.490272)(1790.62,1.436531)(1804.14,1.480876)(1818.98,1.473288)(1832.16,1.434698)(1844.38,1.468459)(1856.83,1.425387)(1870.22,1.483781)(1882.7,1.451498)(1895.07,1.413343)(1908.22,1.422891)(1921.71,1.443326)(1937.24,1.467119)(1949.2,1.426403)(1960.56,1.465204)(1971.97,1.485015)(1983.62,1.436956)(1994.85,1.457637)(2006.42,1.418683)(2018.01,1.443874)(2029.68,1.45654)(2041.49,1.475292)(2053.07,1.410954)(2064.34,1.420016)(2075.7,1.429185)(2087.72,1.462978)(2099.86,1.464637)(2111.41,1.442344)(2123.16,1.467399)(2134.05,1.49004)(2142.21,1.417558)(2151.12,1.425862)(2160.3,1.444847)(2168.7,1.460305)(2176.94,1.468938)(2185.12,1.404847)(2193.33,1.42367)(2201.79,1.439629)(2211.19,1.465629)(2222.18,1.483645)(2231.99,1.435776)(2241.04,1.444049)(2250.51,1.477465)(2260.03,1.409396)(2269.58,1.43528)(2279.25,1.462772)(2289.46,1.484982)(2300.72,1.419709)(2311.6,1.44614)(2322.18,1.466004)(2333.25,1.426205)(2342.14,1.429576)(2350.1,1.447289)(2358.95,1.469231)(2368.02,1.489012)(2377.35,1.444962)(2386.71,1.452906)(2398.03,1.403829)(2408.02,1.437771)(2417.86,1.455889)(2428.43,1.483474)(2438.92,1.428627)(2449.83,1.470186)(2462.03,1.41861)(2473.37,1.451621)(2484.43,1.472704)(2495.06,1.428229)(2503.93,1.436097)(2513.06,1.462592)(2523.04,1.414681)(2531.83,1.43374)(2541.99,1.454313)(2551.51,1.47824)(2560.82,1.488114)(2570.12,1.425993)(2579.27,1.446374)(2589.12,1.475435)(2598.99,1.415756)(2609.07,1.457616)(2620.45,1.488702)(2631.87,1.426012)(2641.58,1.454835)(2650.12,1.475792)(2659.24,1.421637)(2667.45,1.451234)(2675.92,1.46797)(2685.01,1.412433)(2695.4,1.442626)(2705.68,1.473214)(2716.17,1.431659)(2726.92,1.464887)(2737.69,1.404612)(2748.55,1.440463)(2766.18,1.472551)(2779.65,1.487153)(2792.72,1.481225)(2807.39,1.416377)(2822.13,1.448639)(2838.57,1.484628)(2857.49,1.419642)(2873.83,1.430708)(2888.92,1.457306)(2902.57,1.457757)(2910.56,1.476597)(2918.86,1.421283)(2927.14,1.440232)(2935.56,1.452148)(2944.18,1.468816)(2953.91,1.41571)(2963.03,1.4382)(2972.44,1.476348)(2982.75,1.428054)(2993.26,1.45127)(3002.46,1.477203)(3010.62,1.417652)(3018.72,1.43713)(3027.34,1.465749)(3036.62,1.476277)(3046.35,1.420565)(3056.09,1.44029)(3066.4,1.456542)(3077.27,1.409592)(3092.81,1.479847)(3107.26,1.436628)(3120.5,1.464642)(3133.56,1.479157)(3146.39,1.403202)(3159.8,1.443467)(3171.08,1.452815)(3179.94,1.469713)(3188.72,1.486989)(3197.48,1.418625)(3205.95,1.434792)(3214.4,1.452974)(3223.07,1.467336)(3231.98,1.489085)(3241.23,1.425065)(3250.31,1.439027)(3259.63,1.465461)(3269.29,1.481623)(3278.94,1.410316)(3288.63,1.43873)(3298.74,1.481092)(3308.64,1.43018)(3316.93,1.458986)(3325.52,1.482944)(3334.18,1.412542)(3342.65,1.436096)(3351.66,1.46366)(3360.89,1.482735)(3370.59,1.424915)(3380.16,1.448011)(3390.07,1.482983)(3400.24,1.425195)(3410.47,1.43686)(3421.01,1.443752)(3431.43,1.450226)(3440.99,1.467247)(3450.99,1.450171)(3461.26,1.455312)(3472.11,1.428804)(3483.63,1.426815)(3496.99,1.453406)(3510.6,1.448702)(3525.85,1.441301)(3606.26,1.463924)(3655.2,1.433334)(3666.87,1.420633)(3677.29,1.415322)(3688.27,1.446931)(3699.65,1.434828)(3710.97,1.465014)(3723.71,1.460938)(3737.49,1.420135)(3751.69,1.424499)(3766.47,1.433874)(3781.51,1.454822)(3835.11,1.419336)(3850.25,1.417389)(3866.97,1.419582)(3881.22,1.428227)(3896.69,1.440572)(3930.48,1.442834)(3970.38,1.424901)(4003.24,1.484297)(4041.88,1.424148)(4055.59,1.458367)
}; \addlegendentry{Active Gaussians}

\addplot[dashed, black, thick,forget plot] coordinates {(0,1.5) (3500,1.5)};

\addplot[smooth,orange,line width=2pt]
  coordinates{
    (17.336,0.21)(23.655,0.28)(30.651,0.36)(37.187,0.44)(44.164,0.52)(51.227,0.61)(58.681,0.7)(65.294,0.77)(71.871,0.85)(78.969,0.94)(86.445,1.03)(94.046,1.12)(101.331,1.21)(108.928,1.3)(117.28,1.4)(125.847,1.5)(132.737,1.58)(139.621,1.66)(145.8,1.74)(151.969,1.8)(158.651,1.84)(166.044,1.88)(172.849,1.96)(181.123,2.06)(189.151,2.15)(197.328,2.25)(204.902,2.34)(212.604,2.43)(220.714,2.53)(228.913,2.63)(237.699,2.72)(246.038,2.82)(254.367,2.92)(262.747,3.02)(271.259,3.12)(280.232,3.23)(289.169,3.33)(298.363,3.44)(307.903,3.56)(317.023,3.67)(326.317,3.78)(334.75,3.88)(343.127,3.98)(350.932,4)(358.515,4)(365.678,4)(373.398,4)(381.581,4)(389.885,4)(397.898,4)(407.137,4)(415.325,4)(423.533,4)(432.649,4)(441.135,4)(449.639,4)(458.819,4)(467.038,4)(475.391,4)(484.018,4)(492.85,4)(502.224,4)(511.679,4)(520.821,4)(529.959,4)(539.199,4)(548.517,4)(558.748,4)(569.198,4)(580.001,4)(591.912,4)(603.067,4)(614.468,4)(625.939,4)(635.994,4)(644.307,4)(652.905,4)(662.365,4)(672.551,4)(681.858,4)(691.337,4)(700.784,4)(711.571,4)(721.658,4)(732.288,4)(743.134,4)(753.667,4)(762.295,4)(770.198,4)(778.309,4)(788.879,4)(799.672,4)(810.168,4)(819.412,4)(827.341,4)(835.596,4)(844.125,4)(852.528,4)(860.472,4)(868.516,4)(876.565,4)(884.565,4)(893.034,4)(901.481,4)(910.18,4)(919.39,4)(928.592,4)(938.694,4)(950.471,4)(961.806,4)(973.024,4)(983.983,4)(995.079,4)(1006.18,4)(1017.38,4)(1028.82,4)(1040.15,4)(1051.73,4)(1063.63,4)(1075.35,4)(1086.99,4)(1098.56,4)(1110,4)(1121.79,4)(1133.59,4)(1145.21,4)(1153.44,4)(1161.69,4)(1170.13,4)(1178.99,4)(1188.3,4)(1197.79,4)(1207.27,4)(1217.25,4)(1227.2,4)(1237.12,4)(1246.98,4)(1256.9,4)(1266.9,4)(1277.54,4)(1288.01,4)(1298.63,4)(1307.38,4)(1315.68,4)(1324.35,4)(1333.21,4)(1342.3,4)(1351.74,4)(1361.35,4)(1371.04,4)(1380.92,4)(1391.34,4)(1402.48,4)(1413.22,4)(1423.9,4)(1437.23,4)(1448.86,4)(1460.72,4)(1472.33,4)(1483.23,4)(1493.3,4)(1503.26,4)(1513.61,4)(1523.23,4)(1531.34,4)(1539.95,4)(1548.82,4)(1557.98,4)(1567.94,4)(1578.51,4)(1589.26,4)(1597.83,4)(1606.13,4)(1616.71,4)(1627.08,4)(1639.22,4)(1652.66,4)(1666.25,4)(1680.28,4)(1692.84,4)(1705.43,4)(1717.53,4)(1730.51,4)(1741.81,4)(1754.3,4)(1765.85,4)(1778.24,4)(1790.62,4)(1804.14,4)(1818.98,4)(1832.16,4)(1844.38,4)(1856.83,4)(1870.22,4)(1882.7,4)(1895.07,4)(1908.22,4)(1921.71,4)(1937.24,4)(1949.2,4)(1960.56,4)(1971.97,4)(1983.62,4)(1994.85,4)(2006.42,4)(2018.01,4)(2029.68,4)(2041.49,4)(2053.07,4)(2064.34,4)(2075.7,4)(2087.72,4)(2099.86,4)(2111.41,4)(2123.16,4)(2134.05,4)(2142.21,4)(2151.12,4)(2160.3,4)(2168.7,4)(2176.94,4)(2185.12,4)(2193.33,4)(2201.79,4)(2211.19,4)(2222.18,4)(2231.99,4)(2241.04,4)(2250.51,4)(2260.03,4)(2269.58,4)(2279.25,4)(2289.46,4)(2300.72,4)(2311.6,4)(2322.18,4)(2333.25,4)(2342.14,4)(2350.1,4)(2358.95,4)(2368.02,4)(2377.35,4)(2386.71,4)(2398.03,4)(2408.02,4)(2417.86,4)(2428.43,4)(2438.92,4)(2449.83,4)(2462.03,4)(2473.37,4)(2484.43,4)(2495.06,4)(2503.93,4)(2513.06,4)(2523.04,4)(2531.83,4)(2541.99,4)(2551.51,4)(2560.82,4)(2570.12,4)(2579.27,4)(2589.12,4)(2598.99,4)(2609.07,4)(2620.45,4)(2631.87,4)(2641.58,4)(2650.12,4)(2659.24,4)(2667.45,4)(2675.92,4)(2685.01,4)(2695.4,4)(2705.68,4)(2716.17,4)(2726.92,4)(2737.69,4)(2748.55,4)(2766.18,4)(2779.65,4)(2792.72,4)(2807.39,4)(2822.13,4)(2838.57,4)(2857.49,4)(2873.83,4)(2888.92,4)(2902.57,4)(2910.56,4)(2918.86,4)(2927.14,4)(2935.56,4)(2944.18,4)(2953.91,4)(2963.03,4)(2972.44,4)(2982.75,4)(2993.26,4)(3002.46,4)(3010.62,4)(3018.72,4)(3027.34,4)(3036.62,4)(3046.35,4)(3056.09,4)(3066.4,4)(3077.27,4)(3092.81,4)(3107.26,4)(3120.5,4)(3133.56,4)(3146.39,4)(3159.8,4)(3171.08,4)(3179.94,4)(3188.72,4)(3197.48,4)(3205.95,4)(3214.4,4)(3223.07,4)(3231.98,4)(3241.23,4)(3250.31,4)(3259.63,4)(3269.29,4)(3278.94,4)(3288.63,4)(3298.74,4)(3308.64,4)(3316.93,4)(3325.52,4)(3334.18,4)(3342.65,4)(3351.66,4)(3360.89,4)(3370.59,4)(3380.16,4)(3390.07,4)(3400.24,4)(3410.47,4)(3421.01,4)(3431.43,4)(3440.99,4)(3450.99,4)(3461.26,4)(3472.11,4)(3483.63,4)(3496.99,4)(3510.6,4)(3525.85,4)(3606.26,4)(3655.2,4)(3666.87,4)(3677.29,4)(3688.27,4)(3699.65,4)(3710.97,4)(3723.71,4)(3737.49,4)(3751.69,4)(3766.47,4)(3781.51,4)(3835.11,4)(3850.25,4)(3866.97,4)(3881.22,4)(3896.69,4)(3930.48,4)(3970.38,4)(4003.24,4)(4041.88,4)(4055.59,4)
}; \addlegendentry{Active Keyframes}

\addplot[dashed, black, thick, forget plot] coordinates {(0,4.0) (3500,4.0)};

\end{axis}

\end{tikzpicture}
\caption{\textbf{Active Gaussians and Keyframes vs \ac{vram} usage on KITTI Scene 02.}
Our disk-based system maintains constant VRAM usage as scene complexity grows. Active Gaussians plateau at 1.5 M (lower dashed line) and keyframes at 400 (upper dashed line), while the total scene representation continues expanding. Superior quality stems from using more Gaussians overall via disk storage, not from improved Gaussian efficiency.}
\label{fig:vram_usage}
\end{figure}

\begin{table}[tb]
\caption{\textbf{Results on Nvidia Jetson. ATE in cm.}}
\footnotesize
\setlength{\tabcolsep}{4.5pt}
    \centering
    \begin{tabular}{l|cc|cc|cc}
        \toprule
        & \multicolumn{2}{c|}{\textbf{TUM}} & \multicolumn{2}{c|}{\textbf{Replica}} & \multicolumn{2}{c}{\textbf{KITTI}} \\
        \textbf{Metrics} & Desktop & \cellcolor{gray!15}Jetson & Desktop & \cellcolor{gray!15}Jetson & Desktop & \cellcolor{gray!15}Jetson \\
        \midrule
        ATE $\downarrow$ & 1.88 & \cellcolor{gray!15}4.09 & 0.33 & \cellcolor{gray!15}0.33 & 262.54 & \cellcolor{gray!15}239.23 \\
        
        PSNR $\uparrow$ & 21.21 & \cellcolor{gray!15}20.48 & 34.59 & \cellcolor{gray!15}32.73 & 20.89 & \cellcolor{gray!15}19.10 \\
        
        SSIM $\uparrow$ & 0.75 & \cellcolor{gray!15}0.74 & 0.94 & \cellcolor{gray!15}0.93 & 0.73 & \cellcolor{gray!15}0.67\\
        
        LPIPS $\downarrow$ & 0.22 & \cellcolor{gray!15}0.24 & 0.05 & \cellcolor{gray!15}0.07 & 0.27 & \cellcolor{gray!15}0.33 \\
        
        FPS $\uparrow$ & 25.09 & \cellcolor{gray!15}13.19 & 32.41 & \cellcolor{gray!15}12.29 & 1.37 & \cellcolor{gray!15}1.13 \\
        \bottomrule
    \end{tabular}
  \label{tab:jetson_results}
\end{table} 
\section{Limitations \& Future Work}
DiskChunGS scales \ac{3dgs} \ac{slam} through disk-based Gaussian storage but, like other \ac{3dgs} methods, does not handle artifacts such as motion blur, lens flare, or dynamic objects. 

Our locality-focused keyframe strategy minimizes I/O by avoiding excessive chunk swapping, enabling efficient scaling. However, regions visible to only a subset of keyframes may receive fewer optimization constraints than methods with global keyframe access, which can occasionally cause artifacts in transitions. Future work could leverage GPU Direct Storage to reduce chunk-swapping costs and enable more flexible optimization strategies. Additionally, integrating a \ac{3dgs} \ac{lod} system would address current limitations in rendering distant objects, and evaluation on ultra-long sequences (10 km+) would further demonstrate scalability.

\section{Conclusion}
We present DiskChunGS, an out-of-core 3DGS SLAM system that overcomes memory limitations through spatial chunking and disk-based management. By treating large-scale reconstruction as an algorithmic rather than hardware challenge, our method scales to multi-kilometer environments on standard GPUs where previous approaches fail. Evaluations demonstrate superior visual quality across diverse scenarios, from indoor spaces to urban driving sequences. Validated on resource-constrained platforms such as the Jetson Orin and integrated with ROS, DiskChunGS enables practical, photorealistic large-scale 3D reconstruction for real-world robotics. 

\bibliographystyle{IEEEtran}
\bibliography{references}
\end{document}